\documentclass[letterpaper, 10 pt, conference]{ieeeconf}

\IEEEoverridecommandlockouts %
\usepackage{times}

\usepackage{amsmath,amssymb}
\usepackage{amsfonts}
\usepackage{graphicx}
\usepackage{textcomp}
\usepackage{xcolor}
\usepackage{soul}
\usepackage{titletoc}
\usepackage{float}
\usepackage{listings}
\usepackage{multirow}
\usepackage{xparse}
\usepackage{optidef}
\usepackage{algorithm}
\usepackage{algpseudocode}
\usepackage{wrapfig}
\usepackage{url}

\usepackage{multicol}
\usepackage{booktabs}
\usepackage{bm}
\usepackage{overpic}

\usepackage[normalem]{ulem} %

\definecolor{code-constant}{HTML}{d86001}
\definecolor{code-grey}{HTML}{848482} %

\usepackage[noadjust]{cite}

\usepackage{color}
\definecolor{international_orange}{RGB}{240, 74, 0}
\definecolor{citecolor}{RGB}{240, 74, 0}
\usepackage[pagebackref=true,breaklinks=true,urlcolor=citecolor,colorlinks,citecolor=citecolor,bookmarks=false,linkcolor=citecolor]{hyperref}

\newcommand{\updated}[1]{\textcolor{black}{#1}}

\newcommand{\obj}[1]{#1}

\NewDocumentCommand \prop {>{\SplitList{ }} m} {\proposition #1}
\NewDocumentCommand \proposition {g g g g} {\texttt{#1}(#2
  \IfValueTF{#3}{,\,#3}{}
  \IfValueTF{#4}{,\,#4}{}
  )
}

\NewDocumentCommand \actioncall {g g g g} {\text{#1}(
  \IfValueTF{#2}{#2}{}
  \IfValueTF{#3}{,#3}{}
  \IfValueTF{#4}{,#4}{}
  )
}

\DeclareMathOperator*{\argmax}{arg\,max}

\newcommand{\func}[2]{\mathop{}#1\left(#2\right)}
\newcommand{\prob}[1]{\func{p}{#1}}
\newcommand{\given}{\;\middle|\;}

\newcommand{\goal}{\mathcal{G}}
\newcommand{\obs}{\mathbf{o}_1}
\newcommand{\qone}[1]{\func{q_1}{#1}}
\newcommand{\qtwo}[1]{\func{q_2}{#1}}

\newcommand{\qoneinline}{q_1}
\newcommand{\qtwoinline}{q_2}
\newcommand{\encoder}{Enc}
\newcommand{\decoder}{Dec}

\newcommand{\cp}{Points2Plans}
\newcommand{\cpunbold}{Points2Plans}
\newcommand{\cpl}{Points2Plans\textminus Geo}
\newcommand{\eRDl}{eRDTransformer}
\newcommand{\eRDg}{Points2Plans\textminus Delta}
\newcommand{\sampling}{Greedy}
\newcommand{\pairwise}{Pairwise-RD}

\newcommand{\rd}{RD}

\newcommand{\cpwob}{Points2Plans\textminus Feasibility}
\newcommand{\graphs}{Search}

\definecolor{revision_blue}{RGB}{0, 0, 200}

\usepackage{subcaption}
\usepackage[skip=0pt,font=small,labelfont=bf]{caption}
\usepackage{amsmath} 
\newcommand{\shrinka}

\title{Points2Plans: From Point Clouds to Long-Horizon Plans\\with Composable Relational Dynamics}

\begin{document}
\author{%
  Yixuan Huang$^{1,2}$,
  Christopher Agia$^{1}$, 
  Jimmy Wu$^{3}$, 
  Tucker Hermans$^{2,4}$, 
  and Jeannette Bohg$^{1}$ 
  \thanks{$^{1}$%
    Stanford University.
    $^{2}$ University of Utah. 
    $^{3}$ Princeton University. 
    $^{4}$ NVIDIA Research. 
  }%
}
\input{overview_fig}
\maketitle

\begin{abstract}
We present \cpunbold{}, a framework for composable planning with a relational dynamics model that enables robots to solve long-horizon manipulation tasks from partial-view point clouds.
Given a language instruction and a point cloud of the scene, our framework initiates a hierarchical planning procedure, whereby a language model generates a high-level plan and a sampling-based planner produces constraint-satisfying continuous parameters for manipulation primitives sequenced according to the high-level plan. 
Key to our approach is the use of a relational dynamics model as a unifying interface between the continuous and symbolic representations of states and actions, thus facilitating language-driven planning from high-dimensional perceptual input such as point clouds.
Whereas previous relational dynamics models require training on datasets of multi-step manipulation scenarios that align with the intended test scenarios, \cpunbold{} uses only single-step simulated training data while generalizing zero-shot to a variable number of steps during real-world evaluations. %
We evaluate our approach on tasks involving geometric reasoning, multi-object interactions, and occluded object reasoning in both simulated and real-world settings. 
Results demonstrate that \cpunbold{} offers strong generalization to unseen long-horizon tasks in the real world, where it solves over 85\% of evaluated tasks while the next best baseline solves only 50\%.
\end{abstract}

\section{Introduction}\label{sec:introduction}
Before robots can make their entry as general-purpose helpers in e.g., household environments, they must learn to solve \textit{sequential manipulation} tasks in the presence of partial occlusions while receiving high-dimensional sensor data as input.
Consider the ``constrained packing'' task shown in Fig.~1, where the robot must place all cups into the shelf without collision. 
To succeed, the robot has to reason about the long-horizon effects of its actions (e.g. what happens if the first cup is placed at the front of the shelf?) without perfect knowledge of object geometries or poses. 

The most common paradigm for solving sequential manipulation tasks decomposes a task into a sequence of skills for the robot to execute~\cite{ahn2022can, wu2023m}. 
The open problem remains: how to sequence skills without being \textit{myopic}; returning to our example, placing the first cup at the front of the shelf prevents future placements. 
Traditionally, this problem is addressed by task and motion planning (TAMP) systems, which perform a search for feasible solutions at the symbolic and geometric level~\cite{toussaint2015logic, garrett2020pddlstream}. 
However, TAMP typically assumes access to explicit 3D object models and symbolic operators with predefined effects~\cite{garrett2021integrated}; assumptions that may not hold in increasingly unstructured and partially observable environments.
Other approaches leverage policy hierarchies to learn long-horizon strategies with reinforcement learning (RL)~\cite{xu2021deep, dalal2021accelerating, pmlr-v205-shi23a}.
However, these approaches aim to learn skill sequencing strategies for each new long-horizon task, while we seek to compose skills through planning to solve a large set of downstream tasks~\cite{stap, gsc, lin2023text2motion}.
We thereby ask: \textit{How can we enable composable planning in high-dimensional observation spaces without predefined symbolic operators?}

In this paper, we argue that transformer-based relational dynamics (\rd{})~\cite{huang-icra2023-graph-relations, huang-erd-tro2024} is key to enabling composable, long-horizon planning directly from partial-view point clouds.
\rd{} models implicitly capture the symbolic and geometric effects of robot actions in a shared (object-centric) latent space, which facilitates goal-directed planning.
We first propose an \rd{} model architecture that requires only randomized single-step environment transitions $(s, a, s')$ for training, but can be iteratively applied to predict long-horizon trajectories $(s_1, a_1, \ldots, s_H)$ at plan time. 
Each single-step transition $(s, a, s')$ corresponds to the states before and after executing a manipulation primitive (e.g., picking an apple from a basket and placing it on the table), and thus represents an abstraction over low-level trajectories executed on the robot~\cite{pmlr-v205-silver23a}.
Second, we introduce a sampling-based planning algorithm that selects robot actions that maximize the likelihood of symbolic goals predicted by the \rd{} model.
This algorithm uses a new rollout strategy that interweaves \textit{delta-state} prediction of objects in the latent space with object pose updates in geometric space, resulting in greater accuracy over long-horizons compared to predicting absolute object states as in prior work~\cite{huang-erd-tro2024}.
Finally, we leverage large language models~(LLMs)~\cite{bommasani2021opportunities} to accelerate our planner by predicting candidate plan skeletons; in effect, significantly reducing the number of discrete skill sequences our planner must search through for any given task.
The combination of these components forms the \cp{} planning framework.

Our contributions are three-fold: 1) A \textbf{relational dynamics model} that excels at long-horizon prediction of point cloud states without the need to train on multi-step data; 2) A \textbf{latent-geometric space dynamics rollout strategy} that significantly increases the horizons over which predicted point cloud states are reliable for planning; 3) A \textbf{planning framework}, \cp{}, that integrates our \rd{} model, rollout strategy, sampling-based planner, and task planner to solve complex long-horizon tasks.
In extensive experiments, we demonstrate that \cp{} generalizes to sequential manipulation tasks involving partial occlusions, long-horizon geometric dependencies, and multi-object interactions in both simulated and real-world settings. 
For qualitative demonstrations of our approach operating on a mobile manipulator platform and supplementary materials, please refer to our project page available at \href{https://sites.google.com/stanford.edu/points2plans}{sites.google.com/stanford.edu/points2plans}.

\section{Related Work}\label{sec:related-work}

A standard approach to solving \textbf{long-horizon manipulation tasks} sequences manipulation \textit{skills}~\cite{argall2009survey, da2012learning, felip2013manipulation, kalashnikov2018scalable, xu2023xskill, structdiffusion2023} according to a high-level plan produced by symbolic planners~\cite{kaelbling2017learning, huang2019continuous, yuan2022sornet, wu2023m, cheng2023league, kumar2024practice}, %
language models~\cite{ahn2022can, huang2022inner, driess2023palm, lin2023text2motion, liang2023code, singh2023progprompt}, or combinations~\cite{xu2019regression, wang2022generalizable, silver2022pddl, liu2023llm+, zha2023distilling}.
Reusability of the underlying skills should, in theory, support generalization to a variety of tasks.
However, existing methods are often limited by myopic planning strategies.
For example, works employing visuomotor skills seldom consider the feasibility of skill sequences and hence evaluate tasks that do not require geometric reasoning~\cite{ahn2022can, wu2023m}.
Conversely, works that optimize skill sequences for geometrically complex tasks often rely on hand-crafted state representations~\cite{stap, lin2023text2motion, gsc}.
Our work jointly addresses these limitations by enabling long-horizon lookahead planning from high-dimensional observations (i.e., 3D point clouds).

Alternative approaches \textbf{leverage policy hierarchies} to solve long-horizon manipulation tasks through options~\cite{sutton1999between, bacon2017option, nachum2018data}, parameterized action Markov decision processes~\cite{masson2016reinforcement, chitnis2020efficient, dalal2021accelerating, nasiriany2022augmenting, fang2022generalization}, model-based RL~\cite{xu2021deep, shah2021value, shi2022skill}, and meta-learning~\cite{xu2018neural, huang2019neural}. 
Skill chaining has also been used to coordinate dependencies among skills in a sequence~\cite{chen2023sequential, lee2021adversarial}.
These approaches attain strong performance within the distribution of tasks they are trained on, but may struggle to generalize to unseen tasks~\cite{stap, xu2021deep}.
Instead of training on long-horizon demonstration data, our approach relies on random single-step environment transitions to train a dynamics model, which is then used to compose skills for entirely new long-horizon tasks.

A number of works \textbf{learn dynamics models} for planning in high-dimensional observation spaces.
Latent space dynamics are used for model-based control~\cite{paxton2021predicting, ebert2018visual, hafner2019learning, hafner2019dream, sundaresan2023learning, li20223d}, but predict state changes at small timescales.
Graph neural networks can predict deformable~\cite{shi2023robocook, shi2023robocraft} and multi-object~\cite{chang2016compositional, battaglia2016interaction, sanchez2018graph, kipf2018neural, driess2023learning, simeonov2021long} dynamics.
Other works generate demonstration data via differentiable simulation to learn skill abstractions for deformable object planning from high-dimensional input~\cite{lin2022diffskill, lin2022planning}.
Several works learn \rd{} models~\cite{huang-icra2023-graph-relations, huang-erd-tro2024, huang-icra204-memory} that operate on 3D point clouds, but they lack \textit{composability} (i.e., to support multi-step planning, they must be trained on multi-step trajectories) and are only demonstrated on tasks requiring up to three consecutive skills. 
Our method extends the task horizon over which \rd{} predictions are reliable and enables the efficient sequencing of unseen skill sequences at test time.

\section{Problem Setup}\label{sec:problem-setup}
\begin{figure*}[ht]
    \centering
    \includegraphics[width=1.99\columnwidth,clip,trim=0mm 0mm 0mm 0mm]{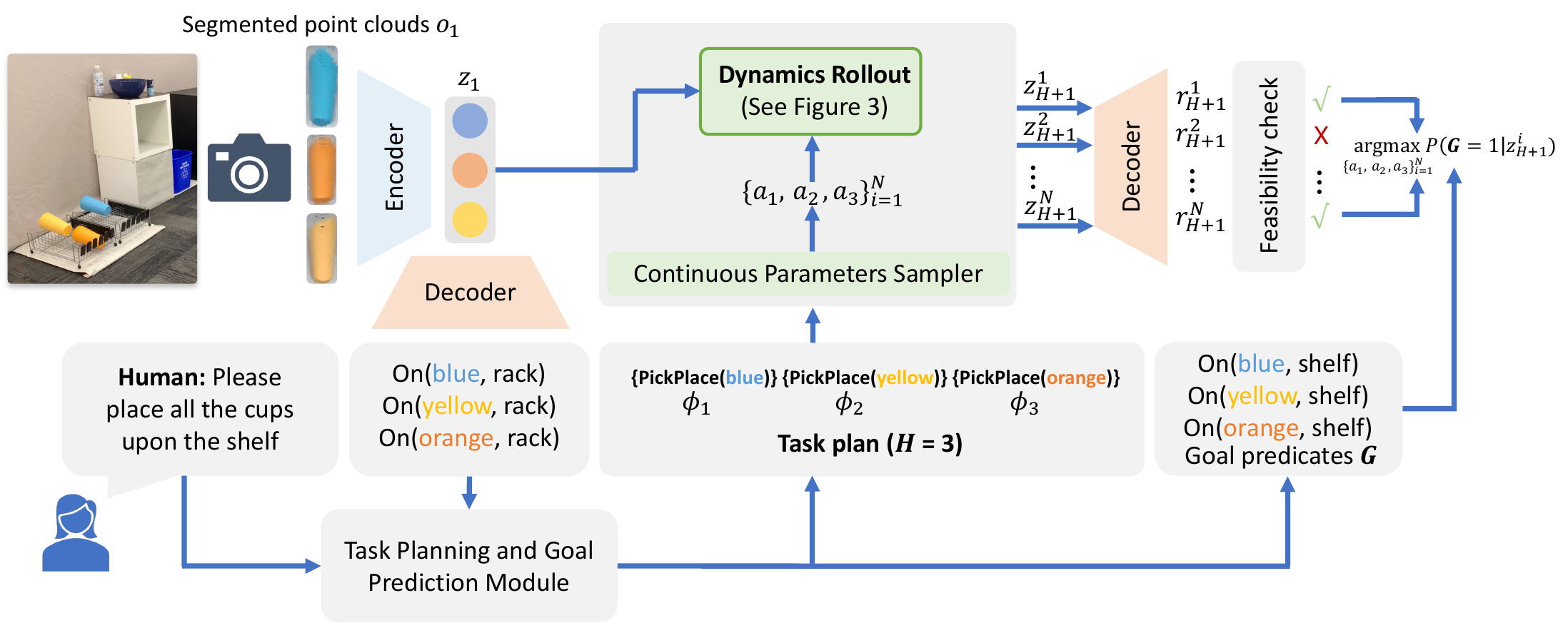} %
    \vspace{5pt}
    \caption{\textbf{Overview of \cp{}.} 
    A partial-view segmented point cloud $\mathbf{o}_1$ is first encoded into the (object-centric) latent state $\mathbf{z}_1$.
    The latent state $\mathbf{z}_1$ is then decoded into predicates that serve as environment context for the task planning and goal prediction module (e.g., an LLM), from which a task plan $\phi_{1:H}$ and a symbolic goal $\mathcal{G}$ are sampled.
    \cp{} then invokes a sampling-based planning procedure to compute continuous parameters $a_{1:H}$ for the manipulation primitives in the task plan $\phi_{1:H}$. 
    Infeasible plans (e.g., collisions) are rejected, and the plan that maximizes the goal likelihood in the final state $\mathbf{z}_{H+1}$ is returned.
    }
    \label{fig:approach}
    \vspace{-15pt}
\end{figure*}

We aim to solve sequential manipulation tasks given segmented, partial-view 3D point clouds of the scene $\mathbf{o}_1$ and a natural language instruction $l$ describing the task.
Satisfying the instruction $l$ entails achieving a goal configuration of objects (i.e., a goal state) $\mathcal{G}$ that can be expressed with predicates from a closed set $\mathcal{R}$.
Predicates are boolean-valued functions that describe object properties e.g., $\prop{Movable a}, \prop{Openable a} \in \mathcal{R}$ and relationships among objects e.g., $\prop{Above a b}, \prop{Inside a b} \in \mathcal{R}$, including those that dictate action feasibility e.g., $\prop{Blocking a b} \in \mathcal{R}$.
A ground predicate is a predicate expressed over specific object instances (e.g., $\prop{Above cup table}$), and a \textit{fact} is an assertion of truth over a ground predicate (e.g., $\prop{Above cup table} = \mathrm{true}$).
Thus, we define the goal state as a conjunction of desired facts $\mathcal{G} = g_1 \land \ldots \land g_M$, where each fact $g_j$ specifies a desired spatial relationship among objects.
In this work, we assume that the closed set of predicates $\mathcal{R}$ is pre-specified, while noting that methods exist to learn predicates from data~\cite{silver2023predicate, shah2024reals}.

\textbf{Manipulation Primitives.}
To solve long-horizon tasks, we assume access to a library of manipulation primitives $\mathcal{L}^\phi = \{\phi^1, \ldots, \phi^K\}$.
Each primitive $\phi^k$ takes as input continuous parameters $a^k \in \mathcal{A}^k$ and executes a trajectory on the robot~\cite{felip2013manipulation}. 
For example, to pick up an object, the parameters $a^{\mathrm{pick}}$ could correspond to a target grasp pose in the object relative frame; instantiating the primitive 
$\func{\phi^{\mathrm{pick}}}{a^{\mathrm{pick}}}$ would move the robot's end-effector to the target pose and close its gripper.
An \textit{action} $\psi^k$ is defined as a pair of a primitive and a parameter $\langle \phi^k, a^k \rangle$.

\textbf{Perception.} 
We assume access to two perception modules: a) a segmentation method that can return segmented point clouds $\mathbf{o}$; b) an object detector that returns the semantic class of each object.
In this work, we use open-source models for segmentation~\cite{kirillov2023segment} and detection~\cite{gu2021open}.

\textbf{The Planning Objective.}
Given an instruction $l$ and segmented partial-view point clouds $\mathbf{o}_1$, our objective is to compute a plan $\tau = [\psi_1, \ldots, \psi_H]$ (we use range subscripts to denote sequences e.g., $\psi_{1:H}$) that when executed  maximizes the probability of the goal implied by instruction $l$:
\begin{equation}\label{eq:planning-objective}
\argmax_{\mathcal{G},\psi_{1:H}}\;\;\;
    \prob{l \given \mathcal{G}, \mathbf{o}_1} 
    \prob{\mathcal{G} \given \psi_{1:H},  \mathbf{o}_1}.
\end{equation}
The first term defines the probability of observing an instruction $l$ given observation $\mathbf{o}_1$ and the user's hidden logical goal $\mathcal{G}$. 
The second term defines the probability of achieving the logical goal $\mathcal{G}$ given the initial observation $\mathbf{o}_1$ and robot actions $\psi_{1:H}$.

\section{Proposed Approach: \cp{}}\label{sec:approach}
The planning objective in Eq.~\ref{eq:planning-objective} can be optimized via a hierarchical approach~\cite{kaelbling2011hierarchical} that first generates a \textit{task plan} in the form of a sequence of primitives $\phi_{1:H}$ and then evaluates its feasibility when planning the parameters $a_{1:H}$ of the primitive sequence.
We formulate our hierarchical planner via two distributions in Eq.~\ref{eq:approximation} 
\begin{equation}\label{eq:approximation}
    \argmax_{G, \psi_{1:H}}
    \qone{\phi_{1:H}, \goal \given l, \obs} \qtwo{a_{1:H} \given \phi_{1:H}, \goal, \obs}.
\end{equation}
The first distribution \(\qoneinline(\phi_{1:H}, \goal | l, \mathbf{o}_1)\) represents the task planner, which serves two roles: a) proposing candidate task plans $\phi_{1:H}$ that are symbolically correct \textit{w.r.t.} the instruction $l$ and initial observation $\mathbf{o}_1$, and b) converting the instruction $l$ into its corresponding goal state $\mathcal{G}$ used to ensure completion of the task.
In this work, we use LLMs~\cite{bommasani2021opportunities} to predict candidate task plans $\phi_{1:H}$ and goals $\mathcal{G}$ from instructions $l$ and textual scene descriptions, while noting that other symbolic~\cite{huang-erd-tro2024} and data-driven~\cite{driess2020deep2} alternatives are possible.

Given a candidate task plan, we must determine whether it can be feasibly executed in the environment and achieve the desired goal.
Therefore, upon sampling $\Tilde{\phi}_{1:H}$ and $\Tilde{G}$ from the task planner $\qoneinline(\phi_{1:H}, \goal | l, \mathbf{o}_1)$, we sample parameters $\Tilde{a}_{1:H}$ from the second distribution $\qtwoinline(a_{1:H} | \Tilde{\phi}_{1:H}, \Tilde{\goal}, \obs)$ to approximately 
solve the optimization problem in Eq.~\ref{eq:approximation}. 
This second distribution represents the probability that parameters $\Tilde{a}_{1:H}$ satisfy the goal \(\Tilde{\mathcal{G}}\) given observation $\mathbf{o}_1$ and task plan $\Tilde{\phi}_{1:H}$.
To obtain parameters $\Tilde{a}_{1:H}$, we propose a long-horizon planning procedure with a transformer-based \rd{} model.
The full planning procedure is visualized in Fig.~\ref{fig:approach}.

In the following sections, we outline our \rd{} model architecture (Sec.~\ref{sec:model-architecture}), a hybrid rollout strategy for predicting point cloud states (Sec.~\ref{sec:hybrid-rollout}), and finally, we present our full planning approach (Sec.~\ref{sec:planning-actions}).

\subsection{Composable Relational Dynamics}\label{sec:model-architecture}
Modeling the effects of actions on the environment (i.e., the \textit{dynamics}) is essential for long-horizon planning. 
Yet, obtaining dynamics models that are both accurate and applicable to a wide range of downstream tasks is challenging for several reasons: a) they are difficult to learn with e.g., imperfect state knowledge or multi-object interactions; b) models trained on one distribution of long-horizon sequences may not generalize well to others. 
To address these challenges, we propose several design considerations for transformer-based RD models~\cite{huang-erd-tro2024} that yield significant improvements in prediction accuracy and allow the model to be chained to predict entirely new long-horizon sequences. 
Our \rd{} model is comprised of three components: an encoder $\encoder$, a transformer-based dynamics model $T$, and a decoder $\decoder$, all of which are jointly trained on single-step environment transitions. 
We describe the details of each component below.

\textbf{Encoder.}
The encoder $\encoder$ takes as input segmented point clouds $\mathbf{o}_t = o_t^1, \ldots, o_t^M$ at timestep $t$ and produces a factored, object-centric latent state $\mathbf{z}_t = z_t^1, \ldots, z_t^M$, where $M$ is the number of objects in the scene (which may vary across tasks).
It embeds each per object segment using PointConv~\cite{wu2019pointconv} and appends a learned positional embedding in PyTorch~\cite{NEURIPS2019_9015} to the resultant per object latent, giving $\mathbf{z}_t = \func{\encoder}{\mathbf{o}_t}$.

\textbf{Dynamics.}
We propose a \textit{delta-dynamics} model $T$ that takes as input the current latent state $\mathbf{z}_t$ and action $\psi_{t} = \langle \phi_{t}, a_t \rangle$, and predicts the \textit{delta state} in the latent space as $\delta\mathbf{z_t} = \func{T}{\mathbf{z}_t, \psi_t}$. 
We use a transformer as the delta-dynamics model $T$ since its inductive bias can represent interactions among the multiple objects in $\mathbf{z}_t$ as a result of action $\psi_{t}$. 
Our hypothesis is that it is easier to learn the relative effect $\delta\mathbf{z_t}$ of an action $\psi_{t}$ than it is to directly predict the resulting \textit{absolute state} $\mathbf{z}_{t+1}$ (i.e., hereafter referred to \textit{absolute dynamics}~\cite{huang-icra2023-graph-relations, huang-erd-tro2024}), since the relative effects of actions might be similar across many states $\mathbf{z}_t$.
We show in Sec.~\ref{sec:experiments} that the choice of delta dynamics translates to notable improvements in pose and predicate prediction accuracy.

\textbf{Decoder.}
The decoder $\decoder$ consists of two heads: a relation decoder $\decoder_r$ and a pose decoder $\decoder_p$.
The relation decoder $\decoder_r$ predicts the probability of each ground predicate being $\mathrm{true}$.
More formally, if $\mathcal{U}$ represents the set of ground predicates, then $\mathbf{r}_t =  \func{\decoder_r}{\mathbf{z}_t} = \{\prob{u=\mathrm{true} \given \mathbf{z}_t} \vert\: \forall u \in \mathcal{U}\}$.
The probability $\prob{u=\mathrm{true} \given \mathbf{z}_t}$ for one ground predicate $u \in \mathcal{U}$ is denoted by $\func{\decoder^u_r}{\mathbf{z}_t}$.
This decoder head can operate on any latent state $\mathbf{z}_t$; for instance, it can be used to detect facts at the initial state as $\mathbf{r}_1 = \func{\decoder_r}{\func{\encoder}{\mathbf{o}_1}} = \func{\decoder_r}{\mathbf{z}_1}$. %
The pose decoder $\decoder_p$ takes as input a delta state in the latent space $\delta\mathbf{z}_t$ and predicts the relative pose change of all objects in the scene as $\delta \mathbf{p}_t = \delta p_t^1, \ldots, \delta p_t^M = \func{\decoder_p}{\delta\mathbf{z}_t}$.
Hence, this decoder can only be applied to delta states predicted by $T$. 

\subsection{Hybrid Rollout Strategy}\label{sec:hybrid-rollout}
\begin{figure}[t]
   \includegraphics[width=0.95\linewidth,clip,trim=0mm 0mm 0mm 0mm]{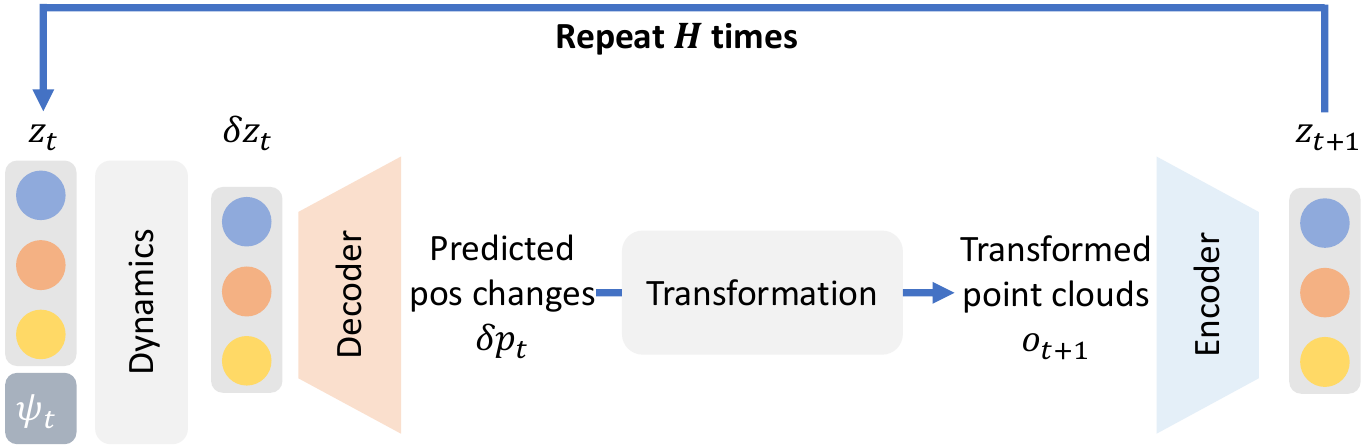} %
   \vspace{5pt}
    \caption{\textbf{\cp{} hybrid rollout strategy.} %
    }
    \label{fig:rollout} 
    \vspace{-15pt}
\end{figure}
We propose a hybrid latent-geometric space dynamics rollout strategy that uses the RD encoder $\encoder$, dynamics $T$, and decoder $\decoder$ (described in Sec.~\ref{sec:model-architecture}) to predict the future states of a given plan $\tau = \psi_{1:H}$, i.e., based on the task plan $\phi_{1:H}$ and its continuous parameters $a_{1:H}$.
The rollout strategy is visualized in Fig.~\ref{fig:rollout}.

Let us consider the first timestep: the hybrid rollout strategy first encodes segmented point cloud $\mathbf{o}_1$ into the latent state $\mathbf{z}_1 = \func{\encoder}{\mathbf{o}_1}$.
Conditioned on the first action $\psi_1$, the delta state is then predicted as $\delta\mathbf{z_1} = \func{T}{\mathbf{z}_1, \psi_1}$. 
The decoder $\decoder_p$ predicts the delta change in pose $\delta \mathbf{p}_1$.
Finally, $\delta \mathbf{p}_1$ is used to transform the point clouds $\mathbf{o}_1$ to obtain $\mathbf{o}_2 = \omega(\delta \mathbf{p}_1)\mathbf{o}_1$.  
This process is repeated for all timesteps $H$ in the plan resulting in the final point cloud $\mathbf{o}_{H+1}$ and latent $\mathbf{z}_{H+1}$ state.
By interweaving latent and geometric state representations, our rollout strategy mitigates compounding prediction errors in the latent space.

\subsection{Planning Action Sequences with Relational Dynamics}\label{sec:planning-actions}
We outline our full approach to planning an action sequence $\psi_{1:H}$ from a language instruction $l$ and the segmented point cloud of the scene $\mathbf{o}_1$ (visualized in Fig.~\ref{fig:approach}). 
Our approach is hierarchical: we solve the optimization problem in Eq.~\ref{eq:approximation} by first generating a candidate task plan $\Tilde{\phi}_{1:H}$ with the LLM and then attempting to sample a set of continuous parameters $\Tilde{a}_{1:H}$ that the robot can feasibly execute.

Given an instruction $l$ with an initial observation $\mathbf{o}_1$, we sample $\Tilde{\phi}_{1:H}$ and $\Tilde{\goal}$ from $\qoneinline(\phi_{1:H}, \mathcal{G} | l, \mathbf{o}_1)$. %
In practice, we use a shooting-based method~\cite{lin2023text2motion}, which queries an LLM few-shot to predict $N$ task plans $\{\Tilde{\phi}^i_{1:H_i}\}_{i=1}^N$ and their corresponding symbolic goals $\{\Tilde{\mathcal{G}}^i\}_{i=1}^N$.
For each task plan $\Tilde{\phi}_{1:H}$ and goal $\Tilde{\mathcal{G}}$ predicted by the LLM, we seek to generate primitive parameters $\Tilde{a}_{1:H}$ from distribution $\qtwoinline(a_{1:H} | \Tilde{\phi}_{1:H}, \Tilde{\goal}, \obs)$ that will approximately maximize the objective in Eq.~\ref{eq:planning-objective}.
We formulate the planning process for parameters $\Tilde{a}_{1:H}$ as the following constrained optimization problem:
\begin{align}\label{eq:constrained-optim}
    \Tilde{a}_{1:H}^* \;=\; \argmax_{\Tilde{a}_{1:H} \sim q_2} \, &\prod_{g \in \Tilde{\mathcal{G}}} \decoder^g_r(\mathbf{z}_{H+1})\\
    \!\!\!\!\!\! \texttt{subject to} \quad 
    & \!\!\!\!\!\! \func{\decoder^c_r}{\mathbf{z}_t} < \epsilon, \; \forall c \in \mathcal{C}, \; \forall t \in 1, \ldots, H+1 \label{eq:constraints} \\
    \texttt{where} \quad & \!\!\!\! \mathbf{z}_t = \func{\encoder}{\mathbf{o}_t}, \; \forall t \in 1, \ldots, H+1 \label{eq:encoder}\\
    & \!\!\!\!\!\! \delta \mathbf{z}_t = \func{T}{\mathbf{z}_t, \langle \Tilde{\phi}_t, \Tilde{a}_t \rangle}, \; \forall t \in 1, \ldots, H \label{eq:dynamics} \\
    & \!\!\!\!\!\! \delta \mathbf{p}_t = \func{\decoder_p}{\delta\mathbf{z}_t}, \; \forall t \in 1, \ldots, H \label{eq:decoder} \\
    & \!\!\!\!\!\!\!\! \mathbf{o}_{t+1} = \omega(\delta \mathbf{p}_t)\mathbf{o}_t, \; \forall t \in 1, \ldots, H\label{eq:transform}
\end{align}
We optimize Eq.~\ref{eq:constrained-optim} to maximize the probability of achieving goal predicates $\Tilde{\mathcal{G}}$ using sampling-based optimization techniques~\cite{rubinstein1999cross}.
The relation decoder $\decoder_r$ is used to compute the probability of a ground goal predicate $g$ holding true in the final latent state $\mathbf{z}_{H+1}$, which we denote with $\decoder^g_r(\mathbf{z}_{H+1})$.
$\mathcal{C}$ in Eq.~\ref{eq:constraints} represents the set of all feasibility-related ground predicates, such as $\prop{Blocking bowl cup}$.
During optimization, we reject parameter sequences $\Tilde{a}_{1:H}$ that violate feasibility constraints, i.e., ground predicates $c \in \mathcal{C}$ whose probability (predicted by the relation decoder $\func{\decoder^c_r}{\mathbf{z}_t}$) exceeds a calibrated threshold $\epsilon_c$.
For example, we would reject a plan that attempts to grasp a $\obj{cup}$ if $\prop{Blocking bowl cup}$ holds $\mathrm{true}$.
The remaining equations~(Eq.~\ref{eq:encoder}-Eq.~\ref{eq:transform}) correspond to the steps of our hybrid rollout strategy (Sec.~\ref{sec:hybrid-rollout}).

For each candidate task plan $\Tilde{\phi}_{1:H}$ and goal $\Tilde{\mathcal{G}}$ predicted by the LLM, we compute their corresponding parameters $\Tilde{a}^*_{1:H}$ via optimization (Eq.~\ref{eq:constrained-optim}). 
If the success probability $\prod_{g \in \Tilde{\mathcal{G}}} \decoder^g_r(\mathbf{z}_{H+1})$ resulting from the optimal plan $\psi^*_{1:H} = \langle \Tilde{\phi}_{1:H}, \Tilde{a}^*_{1:H} \rangle $ exceeds a success threshold $\epsilon_s$ (e.g., $90$\%), we execute the plan on the robot.
However, if no task plan predicted by the LLM is successful or constraint-satisfying, we fall back to a graph search strategy that enumerates all possible primitive sequences up to a specified search depth (as in \cite{huang-erd-tro2024}). This ensures that more task plans will be tested should the LLM fail to produce a correct plan.

\vspace{-5pt}
\section{Experiments}\label{sec:experiments}
We conduct experiments to test the following questions: 
\textbf{Q1:} Can \cp{} generalize to unseen long-horizon tasks despite only being trained on single-step environment transitions? 
\textbf{Q2:} Does our hybrid rollout strategy and delta-dynamics model improve prediction accuracy compared to previous \rd{} rollout formulations?
\textbf{Q3:} Does \cp{} outperform approaches that sequence skills without predicting dynamics or reasoning about feasibility? 
\textbf{Q4:} Can LLMs improve the planning efficiency of \cp{}?
We generate a dataset of over 36,000 random executions of manipulation primitives in IsaacGym~\cite{isaacgym} to train our \rd{} model and use GPT-4~\cite{achiam2023gpt} as the LLM for all experiments. 

\textbf{Dynamics Planning Baselines.}
We test the performance of \cp{} against five planning baselines.
\cpl{} (read ``minus geo") uses the same \rd{} model as \cp{} but performs rollouts exclusively in the latent space, i.e., without the point cloud transformation in our hybrid rollout strategy (Sec.~\ref{sec:hybrid-rollout}).
Conversely, \eRDg{} uses our hybrid rollout strategy but employs an absolute-dynamics model to predict the absolute state $\mathbf{z_t}$ of objects instead of the delta state $\delta\mathbf{z_t}$.
\eRDl{}~\cite{huang-erd-tro2024} represents the current state-of-the-art \rd{} planning approach, which uses a latent space rollout strategy and an absolute-dynamics model.
We train \eRDl{} using single-step transitions instead of multi-step transitions for fair comparison.
\pairwise{}~\cite{paxton2021predicting} is equivalent to \eRDl{} except it only captures pairwise object interactions with a multi-layer perceptron~(MLP) instead of multi-object interactions with transformers.
\sampling{} selects actions $a_{1:H}$ to avoid collisions without dynamics prediction and long-horizon planning.

\textbf{Task Planning Baselines.}
We test the performance of \cp{} under an LLM task planner and two baselines.
\graphs{} performs an exhaustive graph search as in \cite{huang-erd-tro2024} for task planning. 
\cpwob{} uses the LLM without access to the feasibility-related ground predicates, e.g., $\prop{Blocking a b}$.

\begin{figure*}
    \centering
    \includegraphics[width=1.98\columnwidth]{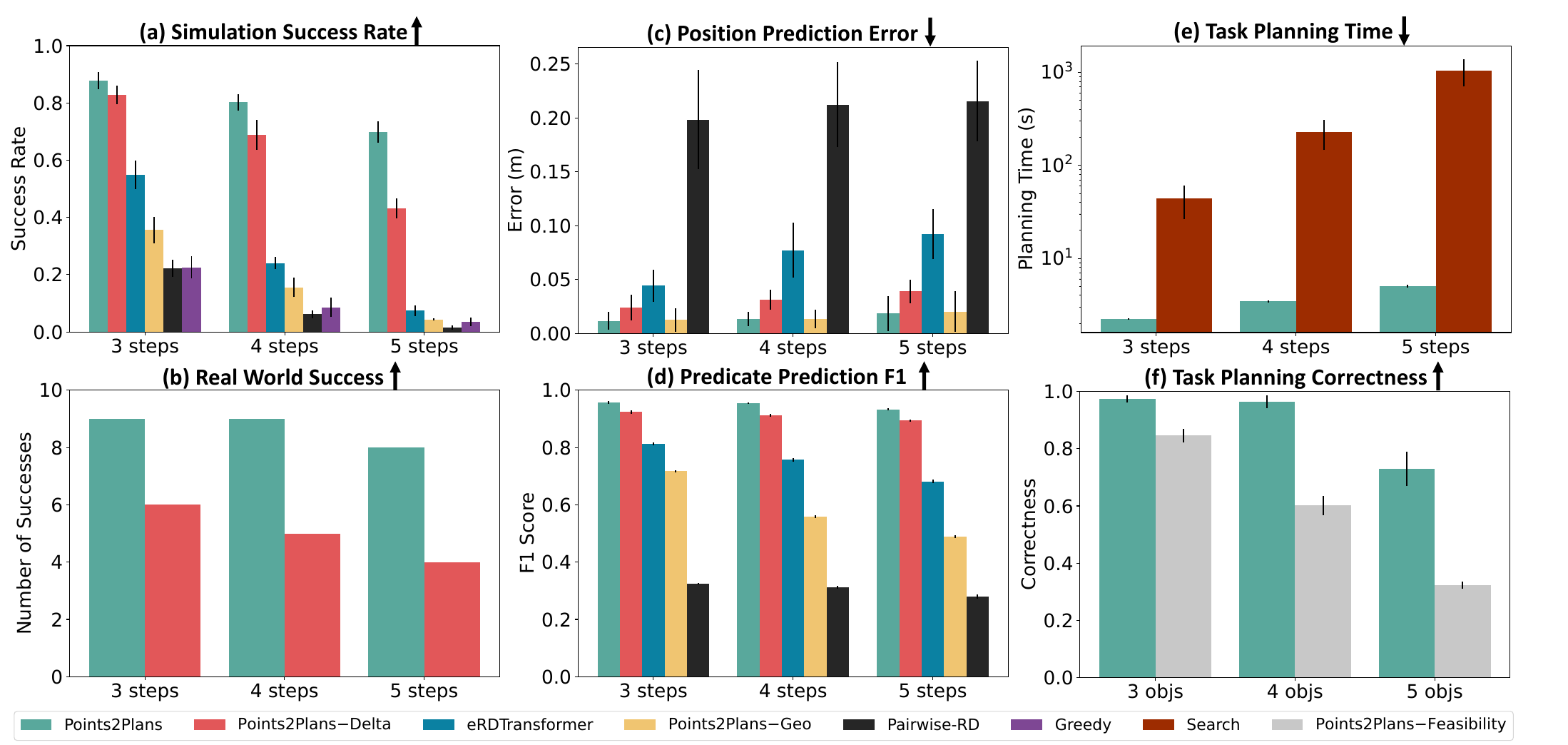}
    \vspace{5pt}
    \caption{\textbf{Simulation and real-world results} for the \textbf{Constrained Packing} (a-d) and \textbf{Constrained Retrieval} (e-f) tasks. 
    As task complexity increases, \cp{} significantly outperforms baselines in terms of planning success rate (a-b), position prediction error (c), and predicate classification accuracy (d).
    Interfacing \cp{} with an LLM task planner increases planning efficiency (e) and correctness (f).
    Planning time is shown on a logarithmic scale. Errors bars denote standard deviations across 500 trials.
    }
    \vspace{-18pt}
    \label{fig:success_comparison}
\end{figure*}

\textbf{Experimental Tasks.}
We evaluate our approach and baselines across a suite of sequential manipulation tasks (Fig.~1).
\textbf{Constrained Packing} tasks the robot with shelving multiple objects in a spatially constrained environment~(e.g., a kitchen cupboard). 
To succeed, the robot must carefully plan the placement positions of the objects so as to avoid collisions. 
We compare \cp{} to the \rd{} baselines on this task as it requires accurate \rd{} predictions of geometric and symbolic states. 
\textbf{Constrained Retrieval} tasks the robot with retrieving target objects in a constrained environment.
To succeed, the robot must identify and remove objects that occlude the target objects before retrieving them.
We compare \cp{} to the task planning baselines on this task as it requires the planner to infer the logically correct task plan based on the initial state.
\textbf{Multi-object Retrieval} tasks the robot with retrieving an object inside a container (e.g., a bowl) in a constrained environment. 
Here, the robot must first remove the container from the constrained environment before grasping the object from inside the container.
This task tests our planner's ability to reason about multi-object interactions and nested geometric dependencies.
\textbf{Occluded Object Retrieval} tasks the robot with retrieving objects in a dark environment (i.e., without perception) given the history of states and actions up until the timestep $t$ at which the lights are turned off.
To succeed, the robot must plan from its \textit{memory} of object positions and relations encoded in the latent state $\mathbf{z}_t$.
We present quantitative results on the \textbf{Constrained Packing} and \textbf{Constrained Retrieval} tasks, and qualitative results on the \textbf{Multi-object Retrieval} and \textbf{Occluded Object Retrieval} tasks.

Across all tasks, we use three manipulation primitives corresponding to pick-and-place, pick-and-toss, and open/close actions.
Further details on our experiments~(e.g., primitives, predicates, hardware, training, prompts, and implementation details) are provided in the supplementary materials made available at \href{https://sites.google.com/stanford.edu/points2plans}{sites.google.com/stanford.edu/points2plans}.
The supplementary materials also includes extended experiments studying Points2Plans' generalization to novel scenarios and robustness to segmentation noise.

\vspace{-5pt}
\section{Results}
\textbf{Simulation Experiments.} We compare \cp{} against all planning baselines
on the \textbf{Constrained Packing} task to evaluate the effect of the \rd{} model and rollout strategy on planning performance.
We run 500 trials for each combination of planning horizon and approach. 
To measure the planning performance, we report success rate, position prediction error, and predicate prediction F1 score.

Results shown in Fig.~\ref{fig:success_comparison}a demonstrate that \cp{} generalizes to unseen long-horizon tasks more effectively than the baselines (Q1).
Comparing \cp{} and \cpl{} in Fig.~\ref{fig:success_comparison}d, we observe that our hybrid rollout strategy contributes greatly to predicate prediction accuracy over long horizons (Q2).
Moreover, comparing \cp{} with \eRDg{}, \eRDl{}, and \pairwise{} in Fig.~\ref{fig:success_comparison}c shows the importance of our delta-dynamics model, as the baselines (which employ an absolute-dynamics model) exhibit a larger accumulation of position prediction error over increasing prediction horizons (Q2).
Finally, we see that the \sampling{} approach performs significantly worse than \cp{} in Fig.~\ref{fig:success_comparison}a, indicating that multi-step planning is required for the long-horizon tasks (Q3). 

\begin{figure}[h]
   \vspace{-5pt}
   \includegraphics[width=0.95\linewidth,clip,trim=0mm 0mm 0mm 0mm]{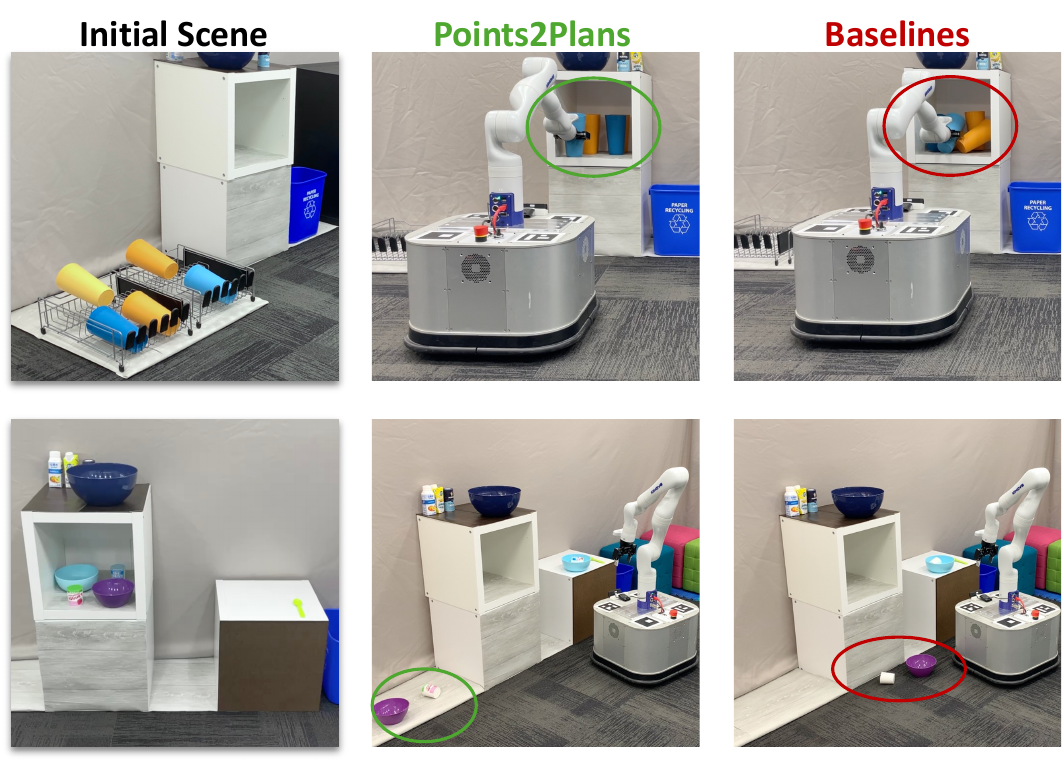} %
   \vspace{5pt}
    \caption{\cp{} generalizes to unseen long-horizon tasks, whereas the baselines struggle to find collision-free plans.}
    \label{fig:baseline_fail} 
    \vspace{-10pt}
\end{figure}

\textbf{Real-World Experiments.} We evaluate \cp{} against \eRDg{} (the next best-performing baseline) in the real world. 
We run 10 trials of each method per task.
The results in Fig.~\ref{fig:success_comparison}b show that \cp{} solves over 85\% of long-horizon tasks and significantly outperforms \eRDg{}, which solves only 50\% of the tasks.
Fig.~\ref{fig:baseline_fail}~(top row) %
illustrates how the baseline fails to plan collision-free placements for multiple objects, due to prediction errors from its \rd{} model.
In contrast, Fig.~1 and Fig.~\ref{fig:baseline_fail} show that \cp{} effectively generalizes to various real-world tasks without fine-tuning (Q1). 

\textbf{Task Planning Ablation.} 
We evaluate the performance of \cp{} when configured with different task planning strategies in the \textbf{Constrained Retrieval} task.
We run 500 trials of each approach per task.
Fig.~\ref{fig:success_comparison}e shows that the planning time of \cp{} increases only linearly with the LLM task planner, whereas the \graphs{} task planner (enumerating all possible discrete parameters) results in an exponential increase in planning time \textit{w.r.t.} to the task horizon (Q4).
In Fig.~\ref{fig:success_comparison}f, we see that the \cpwob{} baseline struggles to predict feasible task plans, highlighting the importance of providing feasibility-related predicates to the LLM task planner as in \cp{}.
Plan executions of \cp{} and \cpwob{} are shown in the bottom row of Fig.~\ref{fig:baseline_fail}.
The baseline fails to remove occluding objects before attempting to grasp the target objects behind them, while \cp{} infers a feasible task plan based on feasibility-related predicates detected by the \rd{} model.

\section{Conclusion}\label{sec:conclusion}

In this work, we study the problem of solving sequential manipulation tasks from partial-view point clouds and language instructions.
We present a long-horizon planning framework, \cp{}, that uses transformer-based relational dynamics to sequence manipulation skills and coordinate their geometric dependencies.
In experiments, we show that interleaving additive, delta-state predictions in the latent space with rigid-body transformations in the geometric space leads to more accurate predictions of point cloud states over long horizons.
As a result, our relational dynamics model can accurately learn the effects of robot skills from a dataset of random, single-step transitions, and then compose the skill effects at planning time to solve multi-step tasks.
We deploy \cp{} on a mobile manipulator platform and demonstrate that it can generalize to diverse real-world tasks such as shelving kitchenware, retrieving occluded objects, and planning from memory.
Future work includes the design of methods to identify and recover from execution failures, online fine-tuning of the relational dynamics model to improve real-world transfer, and interfacing with closed-loop policies to solve tasks that require finer-grained motions.

\section{Acknowledgements}
\noindent{}This work was partially supported by NSF Awards \#2024778 and \#2149585, by DARPA under grant N66001-19-2-4035, and by a Sloan Research Fellowship.
Toyota Research Institute and Toshiba provided funds to support this work.

\bibliographystyle{IEEEtran}
\bibliography{example}

\newpage
\clearpage
\onecolumn

\setlength{\parskip}{1em}

\pagenumbering{arabic}%
\renewcommand*{\thepage}{A\arabic{page}} %
\appendix
\section{Appendix}
\textbf{Overview}

The appendix provides additional details and results. 
First, we present detailed derivations for our planning objective~(Appx.~\ref{sec:problem-deri}) and sampling distributions~(Appx.~\ref{sec:deri}). 
Second, we provide the details of the planning and optimization~(Appx.~\ref{optimization}). 
Third, we include extra experimental details~(Appx.~\ref{sec:predicate} to Appx.~\ref{sec:data_generation}).  
Finally, we provide information on implementation~(Appx.~\ref{sec:NN_details}), failure cases~(Appx.~\ref{sec:failures_analysis}), hardware~(Appx.~\ref{sec:hardware}), generalization experiments~(Appx.~\ref{sec:unseen}), robustness to noisy segmentations~(Appx.~\ref{sec:segmentation}), additional related work~(Appx.~\ref{sec:relatedwork}), and detailed limitations~(Appx.~\ref{sec:limitations}).  
Qualitative results are available at \href{https://sites.google.com/stanford.edu/points2plans}{sites.google.com/stanford.edu/points2plans}. 
\startcontents[sections]
\printcontents[sections]{l}{1}{\setcounter{tocdepth}{2}}
\newpage
\subsection{Generative Model and Problem Formulation}\label{sec:problem-deri}
\begin{figure}[h]
    \centering
    \includegraphics[width=0.89\columnwidth]{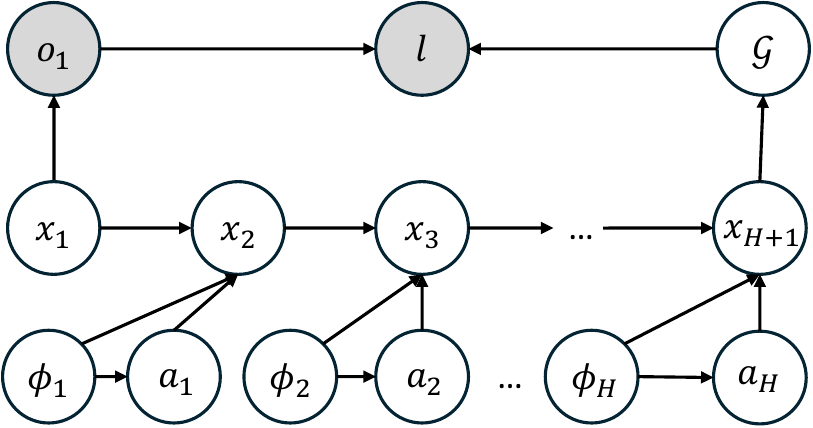}
    \vspace{5pt}
    \caption{A causal Bayes net to derive Eq.~\ref{eq:planning-objective}.
    \(\goal\) represents the goal predicates, $l$ is the language instruction, $o_1$ is the initial observation, $\phi_{1:H}$ are the task plans, $a_{1:H}$ are the continuous parameters, and $\mathbf{x}_{1:H}$ represent world states~(including predicates $\mathbf{r}_{1:H}$ and positions $\mathbf{p}_{1:H}$. 
    Shaded nodes represent observed variables. 
    }
    \label{fig:causal}
\end{figure}
Figure~\ref{fig:causal} shows a causal Bayes net defining the relevant variables to our planning problem. Recall that a Bayesian network defines a factorization via conditional independence over the joint probability distribution of all variables in the model. In our model, the joint distribution is thus, \(
\prob{\goal, \obs, l, \psi_{1:H}, \mathbf{x}_{1:H+1}} = \prob{l \given \goal, \obs}\prob{\goal \given \mathbf{x}_{H+1}}\prob{\obs \given \mathbf{x}_1}\prob{\mathbf{x}_1} \prod_{k=1}^H\prob{\mathbf{x}_{k+1} \given \mathbf{x}_k , a_k, \phi_{k}}\prob{a_k \given \phi_k}\prob{\phi_{k}} 
\).

Here \(\obs\) defines the observed world observation at time step 1. We define the observed language instruction as \(l\) and the unobserved goal predicates as \(\goal\). Our model assumes that the user has a goal in mind that the robot is capable of achieving. However, instead of explicitly writing this goal into a command prompt for the robot, the user provides a natural language command. As such the robot must infer the underlying goal predicates conditioned on the language instruction and the world observation shared between the user and robot.

The variable \(\psi_{1:H}\) defines the robot's plan (sequence of primitives and continuous parameters), while \(\mathbf{x}_{k}\) defines the world state at time \(k\). We represent the state dynamics and action priors in the format common to that used in the planning as inference literature e.g.~\cite{rawlik2012stochastic,conkey2020goal}.

While one could fully separate inference of the goal from the robot planning problem, by treating them as a joint problem we ensure that the robot only infers goals that are feasible for it to achieve. This matches our assumption that the human operator is non-adversarial and providing instructions the robot should be able to perform. The robot's planning task is thus to infer both the goal predicates, \(\goal\) and plan \(\psi_{1:H}\) that achieves this goal. While we are primarily concerned with finding the plan, identifying the goal predicates provides a fixed target for our planner. This goal can also be helpful if we want to replan online to correct for execution errors. As an alternative, we could examine marginalizing out the goal variable by integrating over all possible goals. This would require us to use a form of planning to goal distributions~\cite{conkey2020goal}, which we defer to future work.

Given this model associated with the causal graph in Fig.~\ref{fig:causal}, we can now derive the planning objective of Eq.~\ref{eq:planning-objective} through the following steps. (Note that Eq.~\ref{eq:planning-objective}-Eq.~\ref{eq:transform} appear in the main paper.)
\begin{align}
    & \argmax_{\psi_{1:h}, \goal} \prob{\goal, \obs, l, \psi_{1:H}, \mathbf{x}_{1:H+1}} \\
   & = \argmax_{\psi_{1:H}, \goal} \prob{l \given \goal, \obs}\prob{\goal \given \mathbf{x}_{H+1}}\prob{\obs \given \mathbf{x}_1}\prob{\mathbf{x}_1} \prod_{k=1}^H\prob{\mathbf{x}_{k+1} \given \mathbf{x}_k , a_k, \phi_{k}}\prob{a_k \given \phi_k}\prob{\phi_{k}} \label{eq:model-def}\\
   & = \argmax_{\psi_{1:H}, \goal} \prob{l \given \goal, \obs}\prob{\goal \given \mathbf{x}_{H+1}}\prob{\obs \given \mathbf{x}_1}\prob{\mathbf{x}_1} \prob{\mathbf{x}_{H+1} \given \mathbf{x}_1 , \psi_{1:H}}\prob{\psi_{1:H}} \label{eq:no-prod}\\
   & = \argmax_{\psi_{1:H}, \goal} \prob{l \given \goal, \obs}\prob{\goal \given \mathbf{x}_{H+1}}\prob{\mathbf{x}_1 \given \obs}\prob{\obs} \prob{\mathbf{x}_{H+1} \given \mathbf{x}_1 , \psi_{1:H}} \prob{\psi_{1:H}} \label{eq:obs-cond}\\
   & = \argmax_{\psi_{1:H}, \goal} \prob{l \given \goal, \obs}\prob{\goal \given \mathbf{x}_{H+1}}\prob{\mathbf{x}_{H+1} \given \psi_{1:H}, \obs}\prob{\psi_{1:H}}\prob{\obs} \label{eq:no-init-state}\\ 
   & = \argmax_{\psi_{1:H}, \goal} \prob{l \given \goal, \obs}\prob{\goal \given \mathbf{x}_{H+1}}\prob{\mathbf{x}_{H+1} \given \psi_{1:H}, \obs}\\ 
   & = \argmax_{\psi_{1:H}, \goal} \prob{l \given \goal, \obs}\prob{\goal \given \psi_{1:H}, \obs}\label{eq:final}
\end{align}
The first step is simply applying the factorization of the Bayes net. 
Eq.~\ref{eq:no-prod} makes two changes to remove the product over time steps. First we apply the definition \(\prob{\psi_{1:H}} = \prod_{k=1}^H \prob{a_k\given\phi_k}\prob{\phi_k}\), which we name the plan prior. Second we integrate over (i.e. marginalize out) state trajectories as a function of the actions sequence (i.e. plan) to encode the distribution over terminal states \(\mathbf{x}_{H+1}\) as a function of the initial state and plan.
For Eq.~\ref{eq:obs-cond} we then apply the definition of conditional distributions to the initial state prior and observation function. This allows us to then marginalize out the initial state variable \(\mathbf{x}_1\) in Eq.~\ref{eq:no-init-state}.
The next step removes the prior on the initial observation as it is constant and assumes the prior on plans is uniform and thus also constant. The final step then marginalizes out all possible terminal states to recover our problem as stated in Eq.~\ref{eq:planning-objective}.

\subsection{Approximate Sampling Distributions}\label{sec:deri}
We now turn to the derivation of our approximate sampling distributions $\qone{\phi_{1:H}, \goal, \given l, \obs}$ and 
$\qtwo{a_{1:H} \given \phi_{1:H}, \goal, \obs}$.
We start by taking the first term in Eq.~\ref{eq:planning-objective} and use Bayes' theorem~(Eq.~\ref{eq:bayes}) to condition on the observed language instruction \(l\) 
\begin{align}
    & \argmax_{\psi_{1:H}, \goal} \prob{l \given \goal, \obs}
    \prob{\goal \given \psi_{1:H}, \obs} \label{eq:cond}\\
   & = \argmax_{\psi_{1:H}, \goal}  \frac{\prob{\goal \given l, \obs}\prob{l\given \obs}}{\prob{\goal \given \obs}}
   \prob{\goal \given \psi_{1:H}, \obs}
   \label{eq:bayes}
   \\
    & = \argmax_{\psi_{1:H}, \goal}  \frac{\prob{\goal \given l, \obs}}{\prob{\goal \given \obs}}
   \prob{\goal \given \psi_{1:H}, \obs}
   \label{eq:cond_end}
\end{align}
where we can simplify the numerator in Eq.~\ref{eq:bayes} to Eq.~\ref{eq:cond_end} since the value of \(l\) is known and thus \(\prob{l\given \obs}\) is constant. %
We now turn our attention to the second term in Eq.~\ref{eq:planning-objective} and Eq.~\ref{eq:cond_end}. Here we again use Bayes' theorem and the definitions of conditional probability distributions to rearrange terms from Eq.~\ref{eq:continuous} through Eq.~\ref{eq:rearranging2}:
\begin{align}
   & \argmax_{\psi_{1:H}, \goal}  \frac{\prob{\goal \given l, \obs}}{\prob{\goal \given \obs}}
   \prob{\goal \given \psi_{1:H}, \obs}
 \label{eq:continuous} \\
   & = \argmax_{\psi_{1:H}, \goal}  \frac{\prob{\goal \given l, \obs}}{\prob{\goal \given \obs}}
   \frac{\prob{\psi_{1:H} \given \goal, \obs}\prob{\goal \given \obs}}
   {\prob{\psi_{1:H} \given \obs}}
   \label{eq:rearranging1}\\
   & = \argmax_{\psi_{1:H}, \goal}  \prob{\goal \given l, \obs}
   \frac{\prob{\psi_{1:H} \given \goal, \obs}}
   {\prob{\psi_{1:H} \given \obs}}
   \label{eq:rearranging2}\\
   & = \argmax_{\psi_{1:H}, \goal}
   \prob{\goal \given l, \obs}
   \frac{ \prob{a_{1:H}, \phi_{1:H} \given \goal, \obs}}
   {\prob{a_{1:H}, \phi_{1:H} \given \obs}}
   \label{eq:breaking4} \\
&  = \argmax_{\psi_{1:H}, \goal}
    \prob{\goal \given l, \obs}
    \frac{ \prob{\phi_{1:H} \given \goal, \obs}}
   {\prob{\phi_{1:H} \given \obs}}
    \frac{ \prob{a_{1:H} \given \phi_{1:H}, \goal, \obs}}
   {\prob{a_{1:H} \given \phi_{1:H}, \obs}}
    \label{eq:approx-actions} \\
 & = \argmax_{\psi_{1:H}, \goal} 
    \prob{\goal \given l, \obs}
    \frac{ \prob{\phi_{1:H} \given \goal, l, \obs}}
   {\prob{\phi_{1:H} \given \obs}}
    \frac{ \prob{a_{1:H} \given \phi_{1:H}, \goal, \obs}}
   {\prob{a_{1:H} \given \phi_{1:H}, \obs}}
      \label{eq:add-language} \\
 & = \argmax_{\psi_{1:H}, \goal}
     \prob{\goal \given l, \obs}
     \frac{\prob{\phi_{1:H}, \goal \given l, \obs}}
   { \prob{\goal \given l, \obs}\prob{\phi_{1:H} \given \obs}}
     \frac{ \prob{a_{1:H} \given \phi_{1:H}, \goal, \obs}}
   {\prob{a_{1:H} \given \phi_{1:H}, \obs}}
\label{eq:make-joint} \\
  & = \argmax_{\psi_{1:H}, \goal} 
      \frac{\prob{\phi_{1:H}, \goal \given l, \obs}}
   {\prob{\phi_{1:H} \given \obs}}
       \frac{\prob{a_{1:H}\given \phi_{1:H}, \goal, \obs}}
   {\prob{a_{1:H} \given \phi_{1:H},\obs}}\label{eq:continuous_end}
\end{align}
where Eq.~\ref{eq:breaking4} comes from applying the definition of the action components and Eq.~\ref{eq:approx-actions} comes from the fact that skills \(\phi_{1:H}\) can be chosen before selecting their parameters \(a_{1:H}\). The equality in Eq.~\ref{eq:add-language} results from the existing conditional independence relations allowing us to introduce \(l\) without changing the values of the probability distribution. Eq.~\ref{eq:make-joint} results from the definition of conditional distributions. Finally, this allows us to cancel the term appearing in both the numerator and denominator resulting in Eq.~\ref{eq:continuous_end}. 
These two terms are then the distributions we approximate with our sampling distributions \(\qone{\cdot}\) and \(\qtwo{\cdot}\).

Hence we can now summarize the derivation above as the following relationships
\begin{align}
     \psi_{1:H}^{*}, G^{*} &= \argmax_{\psi_{1:H}, \goal} \prob{l \given \goal, \obs}\prob{\goal \given \psi_{1:H}, \obs} \\
  & = \argmax_{\psi_{1:H}, \goal} 
      \frac{\prob{\phi_{1:H}, \goal \given l, \obs}}
   {\prob{\phi_{1:H} \given \obs}}
       \frac{\prob{a_{1:H}\given \phi_{1:H}, \goal, \obs}}
   {\prob{a_{1:H} \given \phi_{1:H},\obs}}\\
   &  \approx \argmax_{\psi_{1:H}, \goal} 
    \qone{\phi_{1:H}, \goal \given l, \obs} \qtwo{a_{1:H} \given \phi_{1:H}, \goal, \obs} \label{eq:sampling_dists}
\end{align}
where the approximation in Eq.~\ref{eq:sampling_dists} is made by assuming a uniform distribution over actions given the initial observation~(o.e. the denominator becomes constant). 
We can then approximately solve the planning objective in Eq.~\ref{eq:planning-objective} 
by sequentially generating samples from the two approximate sampling distributions as 
$\Tilde{\phi}_{1:H}, \Tilde{G} \sim \qone{\phi_{1:H}, \goal \given l, \obs}$ followed by  
$ \Tilde{a}_{1:H} \sim \qtwo{a_{1:H} \given \Tilde{\phi}_{1:H}, \Tilde{G}, \obs}$.

\subsection{Planning and Optimization Details}\label{optimization}
We use a shooting-based planner to determine the actions $\psi_{1:H}$ from an initial observation $\obs$ and a language instruction $l$. 
We provide the details of the shooting-based \cp{} planner in Alg.~\ref{alg:planner}. 

Note that if no plan predicted by the LLM is successful or constraint-satisfying, we fall back to a search-based strategy that enumerates all possible primitive sequences up to a specified search depth (as in \cite{huang-erd-tro2024}), optimizes them with Eq.~\ref{eq:constrained-optim}, and checks if the plan satisfies any goal predicted by the LLM. (Note that Eq.~\ref{eq:planning-objective}-Eq.~\ref{eq:transform} appear in the main paper.)
In practice, we find that the planner seldom falls back to the search-based strategy; however, it ensures that more primitive sequences will be tested should the LLM fail to produce a correct plan.

Furthermore, we provide more details about the constrained optimization~(Eq.~\ref{eq:constrained-optim}). 
The optimization process includes encoding the point clouds into the latent states~(Eq.~\ref{eq:encoder}), the \textit{delta-state} predictions by the \textit{delta-dynamics}~(Eq.~\ref{eq:dynamics}), decoding the \textit{delta-state} in latent space to relative pose changes with the pose decoder $\decoder_p$~(Eq.~\ref{eq:decoder})
, and point cloud transformations~(Eq.~\ref{eq:transform}).

\begin{algorithm}
\updated{
\small
\caption{Shooting-based \cp{} planner}\label{alg:planner}
\begin{algorithmic}[1]
\State \textbf{globals:} $\textsc{LLM}, \encoder, \decoder^g_r, \decoder^c_r, T, \omega, \qtwo{a_{1:H} \given \phi_{1:H}, \goal, \obs}$ 
\Function{Shooting}{$l, \mathbf{o}_1, C$}
    \State $\{\Tilde{\phi}^i_{1:H_i}\}_{i=1}^N, \{\Tilde{\mathcal{G}}^i\}_{i=1}^N \sim \Call{LLM}{l, \obs}$ \Comment{Generate task plans and goals}
    \For{$i = 1 \ldots N$}
        \State $C = \{\,\}$ \Comment{Initiate candidate set for each task plan and goal}
        \State $\{\Tilde{a}^{j}_{1:H_i}\}_{j=1}^{K} \sim \qtwo{\Tilde{a}_{1:H_i} \given \Tilde{\phi}^i_{1:H_i}, \Tilde{\mathcal{G}}^i, \obs}$ \Comment{Sample actions}
        \For{$j = 1 \ldots K$}
            \State $\mathbf{z}_1 = \func{\encoder}{\mathbf{o}_1}$    \Comment{Encode initial observation}
            \State $\mathbf{z}^j_1 = \mathbf{z}_1$ 
            \State $\mathbf{o}^j_1 = \mathbf{o}_1$ 
            \For{$t = 1 \ldots H_i$}  
                \State $\delta \mathbf{z}^{j}_t = \func{T}{\mathbf{z}^{j}_t, \langle \Tilde{\phi}_t, \Tilde{a}^{j}_t \rangle}$   \Comment{\textit{Delta-dynamics} function}
                \State $\delta \mathbf{p}^{j}_t = \func{\decoder_p}{\delta\mathbf{z}^{j}_t}$ 
                \State $\mathbf{o}^{j}_{t+1} = \omega(\delta \mathbf{p}^{j}_t)\mathbf{o}^{j}_t$ \Comment{Point clouds transformations}
                \State $\mathbf{z}^{j}_{t+1} = \func{\encoder}{\mathbf{o}^{j}_{t+1}}$ \Comment{Encode transformed point clouds}
                \If{$\func{\decoder^c_r}{\mathbf{z}^{j}_{t+1}} >= \epsilon$}
                    \State \textbf{raise} \texttt{collision found, break} \Comment{Collision found, reject this sequence $\Tilde{a}^{j}_{1:H_i}$}
                \EndIf
                \If{$t == H$} \Comment{No collision found}
                    \State $C \gets C \cup \{j\}$ \Comment{Add this sequence to candidate set}
                \EndIf
            \EndFor
        \EndFor
        \If{$C != \emptyset$}
            \State $j^* = \argmax_{j \in C}\; \prod_{g \in \Tilde{\mathcal{G}_i}} \decoder^g_r(\mathbf{z}^{j}_{H_i+1})$
            \State \Return $\Tilde{\phi}^i_{1:H_i}\, \Tilde{a}^{j^*}_{1:H_i}$ \Comment{Return the task plan with the continuous parameters}
        \EndIf
    \EndFor
    \State \textbf{raise} \texttt{LLM failure, fall back to \graphs{}}
\EndFunction
\end{algorithmic}
}
\end{algorithm}

\vspace{-10pt}
\subsection{Predicates Definition}~\label{sec:predicate}
Our system includes unary predicates and binary predicates. 

For unary predicates, our system encodes whether a segment is movable~(e.g., a shelf in a cupboard is not movable while an object on the shelf is movable), whether a segment is a drawer, and whether the drawer is opened or closed. 

For binary predicates, our system includes two kinds. 
First, our system includes nine spatial predicates: \textit{left, right, front, behind, above, below, contact, boundary,} and \textit{inside}. 
The definitions of these predicates are the same as~\cite{huang-icra204-memory}. 
Second, we define feasibility-related predicates to indicate the feasibility of each object. 
We define two feasibility-related predicates: \textit{blocking-behind} and \textit{blocking-inside}. 
We define \textit{blocking-behind}(a, b) as true if behind(b, a), below(a, high-surface), below(b, high-surface), above(a, low-surface), and above(b, low-surface), meaning both a and b are in a constrained environment, and b is behind a. 
We define \textit{blocking-inside}(a, b) as true if inside(b, a), below(a, high-surface), below(b, high-surface), above(a, low-surface), above(b, low-surface), meaning both a and b are in a constrained environment and b is inside a. 

\subsection{LLM Prompt Details}\label{sec:prompts}
Our prompts include prompt templates~(black), LLM output~({\color{code-constant}orange}), and in-context examples~({\color{code-grey}grey}). Placeholders, denoted by braces, are substituted with task-related objects for different scenarios.

The in-context examples are toy examples of the tasks that the LLM solves at test time. 
These toy examples describe the usage semantics of the available primitives and predicates, and help constrain the LLM output.

\vspace{0.3cm}
\noindent\fbox{\parbox{0.99\linewidth}{\small{\texttt{{
I am the reasoning system of a mobile manipulator robot operating in a household environment.
  Given 1) an instruction from a human user and 2) the current symbolic state of the environment, I will
  predict a set of possible symbolic goals that the robot could achieve to fulfill the user's instruction.\\\\
Definitions: \\\\
  - Symbolic states and symbolic goals are defined as a set of predicates expressed over specific objects. \\\\
  - The term 'predicate' refers to an object state (e.g., Opened(cabinet)) or a relationship among objects (e.g., On(cup, shelf)).\\\\
The robot can perceive the following information about the environment:
  - The objects in the environment
  - The states of individual objects
  - The relationships among objects\\\\
The robot can detect the following states of individual objects:
  - Opened(a): Object a is opened
  - Closed(a): Object a is closed\\\\
The robot can detect the following relationships among objects:
  - On(a, b): Object a is on object b
  - Inside(a, b): Object a is in object b\\\\
There may be multiple symbolic goals that fulfill the user's instruction. 
  Therefore, I will format my output in the following form:\\\\
  Goals: List[List[str]]\\\\
  Rules:
  - I will output a set of symbolic goals as a list of lists after 'Goals:'. Each nested list represents one goal
  - I will not output all possible symbolic goals, but the most likely goals that fulfill the user's instruction
  - If there are multiple symbolic goals that fulfill the instruction, I will output the simplest goals first"\\\\
}}}}}
\vspace{0.3cm}

\vspace{0.3cm}
\noindent\fbox{\parbox{0.99\linewidth}{\small{\texttt{{
I am the task planning system of a mobile manipulator robot operating in a household environment.
  Given 1) an instruction from a human user, 2) the current symbolic state of the environment, and 3)
  a set of possible symbolic goals that the robot could achieve to fulfill the user's instruction, I will
  predict a set of task plans that the robot should execute to satisfy the symbolic goals. \\\\ 
Definitions:\\\\
- Symbolic states and symbolic goals are defined as a set of predicates expressed over specific objects. \\\\
- The term 'predicate' refers to an object state (e.g., Opened(cabinet)) or a relationship among objects (e.g., On(cup, shelf)). \\\\
- A task plan is a sequence of actions that the robot can execute (e.g., Pick(cup, table), Place(cup, shelf)) \\\\
The robot can perceive the following information about the environment: \\\\
- The objects in the environment \\\\ 
- The states of individual objects \\\\
- The relationships among objects \\\\
The robot can detect the following states of individual objects: \\\\
- Opened(a): Object a is opened 
- Closed(a): Object a is closed \\\\
The robot can detect the following relationships among objects: \\\\ 
- On(a, b): Object a is on object b 
- Inside(a, b): Object a is in object b \\\\
The robot can execute the following actions: \\\\
- Pick(a, b): The robot picks object a from object b \\\\
- Place(a, b): The robot places object a on or in object b \\\\
- Open(a): The robot opens object a \\\\
- Close(a): The robot closes object a \\\\
Action preconditions: \\\\
- If the robot is already holding an object, it CANNOT Pick, Open, or Close another object \\\\
- The robot CAN ONLY Place an object that it is already holding \\\\
There may be multiple symbolic goals that fulfill the user's instruction. Therefore, I will format my output in the following form: \\\\
Plans: List[List[str]] \\\\
Rules: \\\\
- I will output a set of task plans as a list of lists after 'Plans:'. Each nested list represents one task plan \\\\
- I will output one task plan for each symbolic goal. Hence, each goal and its corresponding plan will be located at the same index in the 'Goals' and 'Plans' lists \\\\
- I will only output task plans that are feasible with respect to the defined action preconditions.
}}}}}
\vspace{0.3cm}

\vspace{0.3cm}
\noindent\fbox{\parbox{0.99\linewidth}{\small{\texttt{{
Instructions: Put all the objects on the shelf\\\\
Objects: ['\{object 1\}', '\{object 2\}', '\{object 3\}', 'ground', 'shelf'] \\\\
Predicates: ['On(\{object 1\}, ground', 'On(\{object 2\}, ground', 'On(\{object 3\}, ground'] \\\\
Goals: {\color{code-constant}['On(\{object 1\}, shelf', 'On(\{object 2\}, shelf', 'On(\{object 3\}, shelf']} \\\\
Plans: {\color{code-constant}['Pick(\{object 1\}, ground)', 'Place(\{object 1\}, shelf)', 'Pick(\{object 2\}, ground)', 'Place(\{object 2\}, shelf)', 'Pick(\{object 3\}, ground)', 'Place(\{object 3\}, shelf)']} }}}}}
\vspace{0.3cm}

\vspace{0.3cm}
\noindent\fbox{\parbox{0.99\linewidth}{\small{\texttt{{
Instructions: Retrieve object 1.\\\\
Objects: ['\{object 1\}', '\{object 2\}', '\{object 3\}', '\{object 4\}', 'ground', 'shelf'] \\\\
Predicates: ['On(\{object 1\}, shelf', 'On(\{object 2\}, shelf', 'On(\{object 3\}, shelf', 'On(\{object 4\}, shelf', 'Blocking(\{object 3\}, \{object 4\})', 'Blocking(\{object 3\}, \{object 1\})', 
'Blocking(\{object 2\}, \{object 1\})', 'Blocking(\{object 4\}, \{object 1\})'] \\\\
Goals: {\color{code-constant}[ '\{On(object 2\}, ground', '\{On(object 3\}, ground', '\{On(object 4\}, ground', '\{On(object 1\}, ground']} \\\\
Plans: {\color{code-constant}['Pick(\{object 3\}, shelf)', 'Place(\{object 3\}, ground)', 'Pick(\{object 2\}, shelf)', 'Place(\{object 2\}, ground)', 'Pick(\{object 4\}, shelf)', 'Place(\{object 4\}, ground)', 'Pick(\{object 1\}, shelf)', 'Place(\{object 1\}, ground)']} }}}}}
\vspace{0.3cm}

\textbf{In-context Examples:}

\vspace{0.3cm}
\noindent\fbox{\parbox{0.99\linewidth}{\small{\texttt{{\color{code-grey}
Instructions: Put object 1 on the sink.\\\\
Objects: ['object 1', 'sink', 'kitchen table'] \\\\
Predicates: ['On(object 1, kitchen table'] \\\\
Goals: ['On(object 1, sink'] \\\\
Plans: ['Pick(object 1, kitchen table)', 'Place(object 1, sink)'] }}}}}
\vspace{0.3cm}

\vspace{0.3cm}
\noindent\fbox{\parbox{0.99\linewidth}{\small{\texttt{{\color{code-grey}
Instructions: Get me object 1 from the drawer. I'm in the bedroom. Don't leave the drawer open.\\\\
Objects: ['object 1', 'object 2', 'drawer', 'closet', 'bed'] \\\\
Predicates: ['Inside(object 1, drawer)', 'Inside(object 2, closet)', 'Closed(drawer)', 'Closed(closet)'] \\\\
Goals: ['On(object 1, bed)', 'Closed(drawer)'] \\\\
Plans: ['Open(drawer)', 'Pick(object 1, drawer)', 'Place(object 1, bed)', 'Close(drawer)'] }}}}}
\vspace{0.3cm}

\vspace{0.3cm}
\noindent\fbox{\parbox{0.99\linewidth}{\small{\texttt{{\color{code-grey}
Instructions: Bring me object 2. I'm sitting on the reading chair by the coffee table.\\\\
Objects: ['object 1', 'object 2', 'bookshelf', 'reading chair', 'coffee table'] \\\\
Predicates: ['On(object 1, object 2', 'On(object 2, bookshelf'] \\\\
Goals: [['On(object 2, coffee table)'], ['On(object 2, reading chair)']] \\\\
Plans: [
  ['Pick(object 1, object 2)', 'Place(object 1, bookshelf)', 'Pick(object 2, bookshelf)', 'Place(object 2, coffee table)'],
  ['Pick(object 1, object 2)', 'Place(object 1, bookshelf)', 'Pick(object 2, bookshelf)', 'Place(object 2, reading chair)']
] }}}}}
\vspace{0.3cm}

\vspace{0.3cm}
\noindent\fbox{\parbox{0.99\linewidth}{\small{\texttt{{\color{code-grey}
Instructions: Please retrieve object 1.\\\\
Objects: ['object 1', 'object 2', 'shelf', 'ground'] \\\\
Predicates: ['On(object 1, shelf', 'On(object 2, shelf'] \\\\
Goals: [['On(object 1, shelf)'], ['On(object 2, shelf)
']] \\\\
Plans: [
  ['Pick(object 1, shelf)', 'Place(object 1, ground)'],
  ['Pick(object 2, shelf)', 'Place(object 2, ground)', 'Pick(object 1, shelf)', 'Place(object 1, ground)',
  ]
] }}}}}
\vspace{0.3cm}

\subsubsection{Connections between \rd{} Models and LLMs}
We have several connections as the interface between \rd{} models and LLMs. 
First, given the predicates \textit{above}(A, B) and \textit{contact}(A, B), then \textit{on}(A, B) holds true, and vice versa. 
Second, given the plans as pick(A, D) and place(A, C) from LLMs, the \rd{} models will receive this plan as pick-and-place(A,C).

\subsection{Dataset Generation and Training Details}~\label{sec:data_generation}
\begin{figure}[h]
    \centering
    \includegraphics[width=0.99\columnwidth]{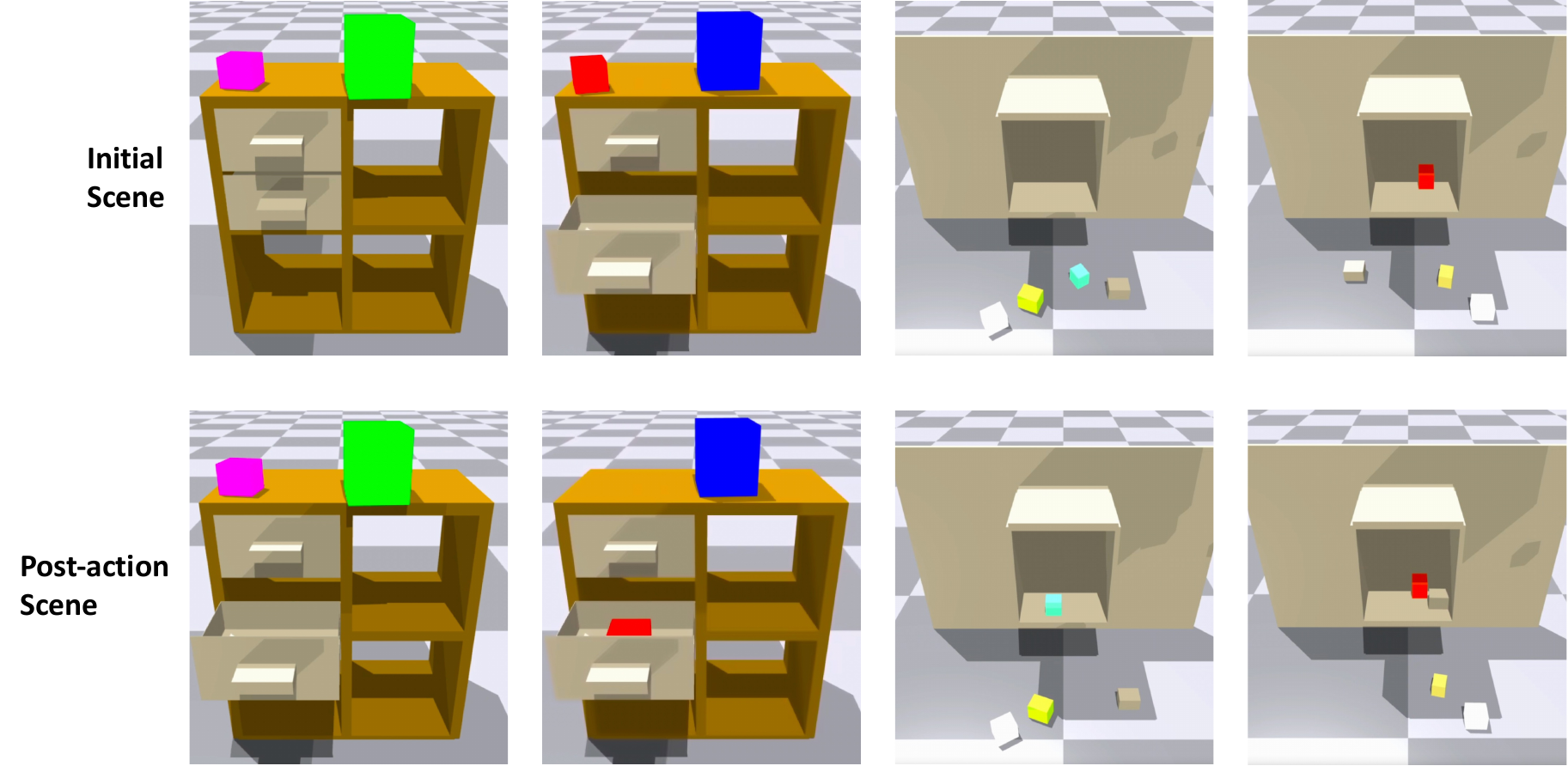}
    \vspace{5pt}
    \caption{We show several examples of single-step primitive executions in simulation. The first two columns show examples of drawers, while the next two columns show examples of the constrained cupboard with different numbers of objects. 
    }
    \label{fig:simulation}
\end{figure}

We generate the training datasets in the IsaacGym~\cite{isaacgym} simulator. 
First, we generate a variable number of randomized objects (size and pose) and save the object pose, segmented point cloud, and predicate as $(\mathbf{o}_t, \hat{\mathbf{r}}_t, \hat{\mathbf{p}}_t)$. 
Then we randomly execute a primitive in the simulator and save the primitive $\psi_t$. 
We teleport the objects to model the effects of each primitive.
After the primitive execution, we record the post-action scene as $(\mathbf{o}_{t+1}, \hat{\mathbf{r}}_{t+1}, \hat{\mathbf{p}}_{t+1})$. 
The dataset contains more than 36,000 primitive executions. 
We show several single-step simulation executions in Fig.~\ref{fig:simulation}. 

We set up one camera in simulation to generate the segmented point clouds. 
Due to the generalization ability of our \rd{} framework to different view points, we can position the real-world camera at different view angles, as long as the object point clouds are within a suitable range to ensure decent quality.
For example, we use the Realsense D435 camera for real-world experiments, with an ideal range of 0.3m to 3m. 
Note that we focus on the critical segments of environmental point clouds~(e.g., horizon surfaces for the cupboard and drawers for the table), as not all segments are visible due to the partial-view nature of the input point clouds. 

We define loss functions with four terms for each transition $(\mathbf{o}_t, \hat{\mathbf{r}}_t, \hat{\mathbf{p}}_t, \psi_t, \mathbf{o}_{t+1}, \hat{\mathbf{r}}_{t+1}, \hat{\mathbf{p}}_{t+1})$ in the training datasets. 
First, we obtain $\mathbf{z}_t = \func{\encoder}{\mathbf{o}_t}$ and $\mathbf{z}_{t+1} = \func{\encoder}{\mathbf{o}_{t+1}}$.
To enable our framework to detect the current predicates, we define the cross-entropy loss between the currently detected predicates and the ground-truth predicates: $L_{cp} = CE(\func{\decoder_r}{\mathbf{z}_t}, \hat{\mathbf{r}}_t) + CE(\func{\decoder_r}{\mathbf{z}_{t+1}}, \hat{\mathbf{r}}_{t+1})$.  
Second, to enable the model to accurately predict the change of pose, we define the second loss term as $L_{pos} = a \cdot \sqrt{b \cdot ||\delta \mathbf{p}_t - ( \hat{\mathbf{p}}_{t+1} - \hat{\mathbf{p}}_{t})||}$. We use two parameters, $a, b$, to balance $L_{pos}$ with other loss terms like $L_{pd}$. 
In practice, we use a = 5 and b = 12. 
Third, to regularize the latent states, we first obtain the predicted latent states as $\mathbf{z}^{\prime}_{t+1} = \mathbf{z}_t + \delta \mathbf{z}_t$. 
We define the regularization loss term as $ L_{reg} =  ||\mathbf{z}_{t+1} - \mathbf{z}^{\prime}_{t+1}||^2_2$. 
Fourth, to predict the future predicates, we define a cross-entropy loss between the predicted predicates and the ground-truth predicates as $L_{fp} = CE(\func{\decoder^q_r}{\mathbf{z}^{\prime}_{t+1}}, \hat{\mathbf{r}}_{t+1})$. 
We train our framework end-to-end with the sum of these four loss terms as $L = L_{cp} + L_{pos} + L_{reg} + L_{fp}$ using the Adam optimizer with a learning rate of $1e-4$. 
We only train our framework with single-step transitions while it can solve long-horizon planning problems in a composable way.

\subsection{Baseline Comparison Details}\label{task}
\begin{figure}[h]
    \centering
    \includegraphics[width=0.99\columnwidth]{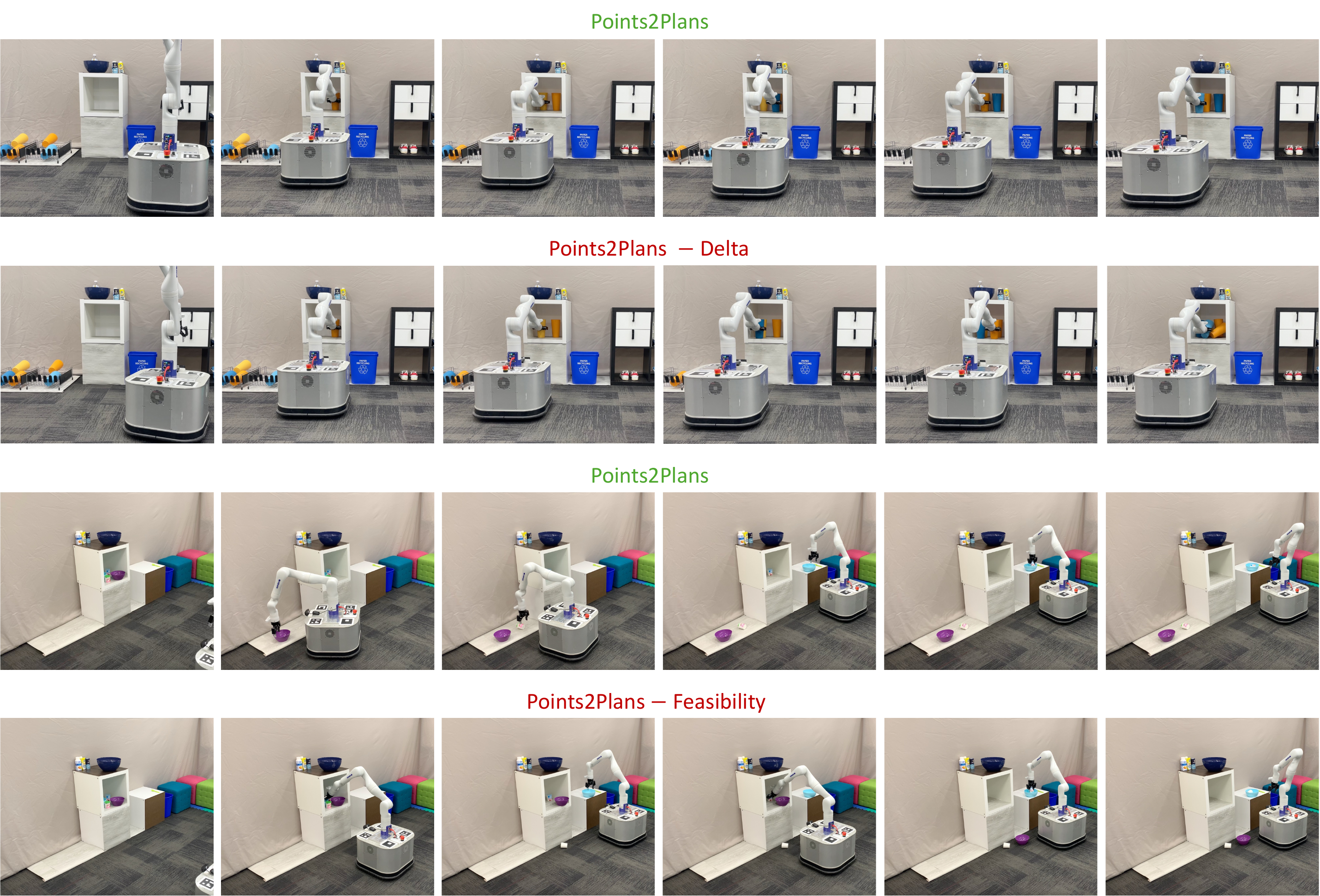}
    \vspace{5pt}
    \caption{We show two failure cases of baselines. The first two rows demonstrate that \cp{} succeeds while \eRDg{} fails in the ``constrained packing'' task. 
    The next two rows show that \cp{} succeeds while \cpwob{} fails in the ``constrained retrieval'' task. 
    }
    \label{fig:baseline_failures}
\end{figure}

We show the details of how baselines fail in the ``constrained packing'' task and the ``constrained retrieval'' task in Fig.~\ref{fig:baseline_failures}.  
Please refer to the supplemental video for the demos.  

\subsection{Primitives Definition}\label{sec:primitives}
We use three primitives in this paper: 
pick-and-place, pick-and-toss, and open/close. 
Since we have a mobile manipulator, we separate the movements of the mobile base and the Kinova arm. 
Based on the objects to manipulate, we first move the mobile base to a reachable space for the arm to manipulate the objects. 
Then, we run the arm planner to manipulate the objects. 

Pick-and-place is defined as first grasping the object and then placing it on the supporting surface. 
For the grasp, we use the point cloud center of each segment to generate the grasps.
We grasp the center of the objects except for large objects. 
For large objects, we use a heuristics offset. For example, we grasp the side of a bowl instead of the center. 

For the placement, we generate a placement height based on the surface height plus a height offset. 
For the toss, we first move the base to a position at a fixed distance from the target, then execute the predefined toss trajectory. 

For the open/close actions, we first determine the handle center using the segmented point clouds. 
Then we move the robot to the pre-open/close position. 
After this, the arm executes the motion with continuous parameters encoding how much the drawer will open or close.

\subsection{Neural Network Implementation Details}~\label{sec:NN_details}
Our \rd{} model is composed of three components: an encoder $\encoder$, a transformer-based dynamics model $T$, and a decoder $\decoder$.
We describe the details of each component below.

\textbf{Encoder:}
We first use the farthest point sampling method to downsample each point cloud to 128 points. 
Based on the input as segmented point clouds $\mathbf{o}_t = o_t^1, \ldots, o_t^M$ at timestep $t$, we first use a PointConv~\cite{wu2019pointconv} to get per object features as $P_t^i = PointConv(o_t^i)$.
The PointConv model we use incorporates three set abstraction layers. 
Each abstraction layer receives input points data and input points position data.
The output from each layer consists of sampled points position data and sampled points feature data, with the input and output points position data having 3 channels. 
The first set abstraction layer has 128 points with 8 samples, utilizing a bandwidth of 0.1. 
It employs an MLP with 3+3 input channels, 32 output channels, and a kernel size of 1.   
The second layer has 64 points with 16 samples and a bandwidth of 0.2, using an MLP with 32+3 input channels, 64 output channels, and a kernel size of 1. 
The third layer is a group\_all layer, generating 128-dimension features per segment with a bandwidth of 0.4, and utilizes an MLP with 64+3 input channels, 128 output channels, and a kernel size of 1. 

Then we concatenate per object feature with the positional embedding of each object in PyTorch~\cite{NEURIPS2019_9015} as $z_t^i = P_t^i \oplus I_i$ where $I_i = Emb_{pos}(i)$. 
Each positional embedding has 128-dimension features. 
Next, we combine the per object latent into an object-centric latent state $\mathbf{z}_t = z_t^1, \ldots, z_t^M$, where each $z_t^i$ has 256 features. 

\textbf{Dynamics:}
The \textit{delta-dynamics} model $T$ takes the input as $\mathbf{z}_t$ and $\psi_{t} = \langle \phi_{t}, a_t \rangle$.
$\phi_{t}$ includes skill id $si$, manipulated obj id $mi$, and placement surface id $pi$. 
For each $si$, we use a different dynamics model $T_{si}$ with a transformer. 
For the transformers,  we utilize 2 sub-encoder layers, 2 heads in the multi-head attention models, and an input/output model size of 256.  

We encode each $a_t$ with an action encoder~($MLP_{si}$) as $a^{m}_t = I_{mi} \oplus MLP_{si}(a_t)$, where $I_{mi}$ encodes which object this primitive will operate on. 
We further use the placement id to represent which surface to place the object on, as $a^{p}_t = I_{pi} \oplus MLP_{si}(a_t)$. 
If there is no surface to place, for example, in an open drawer action, we use zero embeddings for $I_{pi}$. 
The action encoder is a two-layer MLP, with each layer containing 128 neurons. 
Then the dynamics model $T_{si}$ takes the input as M+2 tokens $z_t^1, \ldots, z_t^M, a^{m}_t, a^{p}_t$. 
We discard the action tokens at the output head and obtain the output $\delta\mathbf{z_t} = \delta z_t^1, \ldots, \delta z_t^M$

\textbf{Decoder:}
Based on the latent state $\mathbf{z}_t$, we use different MLPs for different output heads. 
First, we use one MLP for unary predicate prediction: 
$\mathbf{r}^u_t =  \func{\decoder^u_r}{\mathbf{z}_t}$.
Second, we use one MLP for constrained binary predicates prediction: $\mathbf{r}^c_t =  \func{\decoder^c_r}{\mathbf{z}_t}$.
Third, we use one MLP for spatial binary predicate prediction: 
$\mathbf{r}^s_t =  \func{\decoder^s_r}{\mathbf{z}_t}$.

For $\decoder^u_r$, we use a two-layer MLP with a hidden layer of 64 neurons. The output contains 3 bits. 
The first bit encodes whether the segment is a shelf or an object. 
The second and the third bits encode whether the segment is a drawer and whether the drawer is open, respectively. 
For $\decoder^c_r$, we use a three-layer MLP and each hidden layer contains 64 neurons. 
The output contains 2 bits representing \textit{blocking-behind} and \textit{blocking-inside}. 
For $\decoder^s_r$, we use a three-layer MLP with each hidden layer containing 64 neurons. 
The output contains 9 bits representing 9 different spatial predicates. 

The pose decoder $\decoder_p$ takes as input a \textit{delta-state} in the latent space $\delta\mathbf{z}_t$ and predicts the relative pose change of all objects in the scene as $\delta \mathbf{p}_t = \delta p_t^1, \ldots, \delta p_t^M = \func{\decoder_p}{\delta\mathbf{z}_t}$.
Hence, this decoder can only be applied on \textit{delta-state} predicted by $T$. 

For $\decoder_p$, we use a two-layer MLP with one hidden layer containing 64 neurons. The output head contains 2 bits representing $\delta x, \delta y$. 
For $z$, we encode this parameter as part of our discrete parameter for the supporting surface, as shown in the primitive definitions in Sec.~\ref{sec:primitives}. 
LLMs will generate it as part of the task plan.  

For the MLPs, we use Sigmoid in the output head of predicate decoders $\decoder^u_r, \decoder^c_r,$ and $\decoder^s_r$ since these decoders output binary variables. 
For all other MLPs, we use ReLU as the activation function. 

\subsection{Failure Cases Analysis}\label{sec:failures_analysis}
\begin{figure}[h]
    \centering
    \includegraphics[width=0.99\columnwidth]{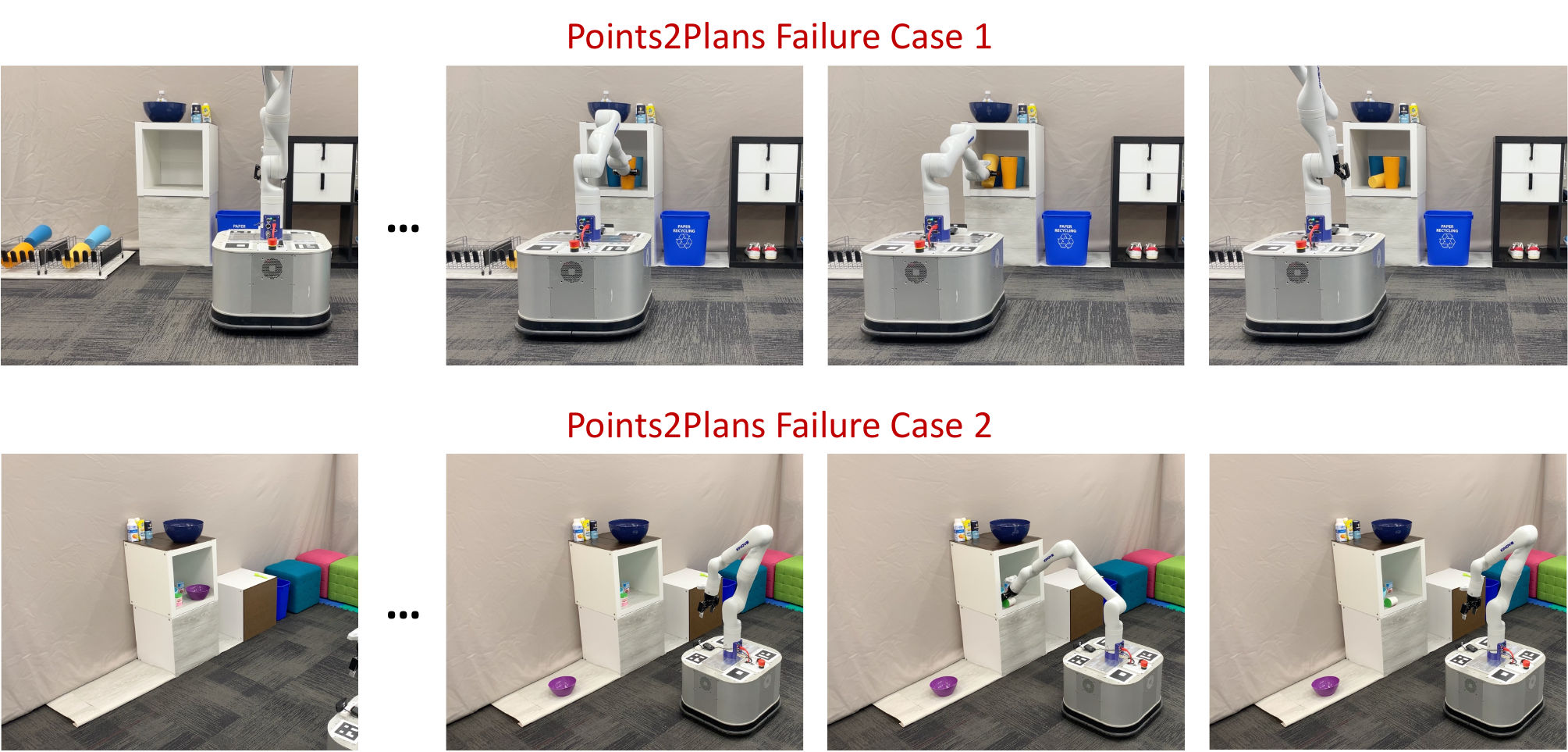}
    \vspace{5pt}
    \caption{Two failure cases of \cp{}. The first row shows a failure case due to an unstable placement in the ``constrained packing'' task. The second row demonstrates that \cp{} fails in the ``constrained retrieval'' task because of a failed grasp. 
    }
    \label{fig:cp_failures}
\end{figure}
We show two failure cases of \cp{} in Fig.~\ref{fig:cp_failures}. These failure cases are caused by primitive execution failures. They highlight the limitations of open-loop execution in \cp{} and motivate the incorporation of closed-loop policies~\cite{chi2023diffusionpolicy, prasad2024consistency} in future work. 
Please refer to our website~(\href{https://sites.google.com/stanford.edu/points2plans}{sites.google.com/stanford.edu/points2plans}) for detailed executions of these two failure cases. 

\subsection{Hardware Setup}~\label{sec:hardware}
The models are trained on a standard workstation with a single GPU~(NVIDIA GeForce RTX 3090 Ti, 24 GB). 
All real-world experiments are conducted on a mobile platform with a custom mobile base and a Kinova arm.
We use a RealSense D435 camera for perception in the real world. 

\subsection{Generalization to Unseen Scenarios}\label{sec:unseen}
\begin{figure}[h]
    \centering
    \includegraphics[width=0.99\columnwidth]{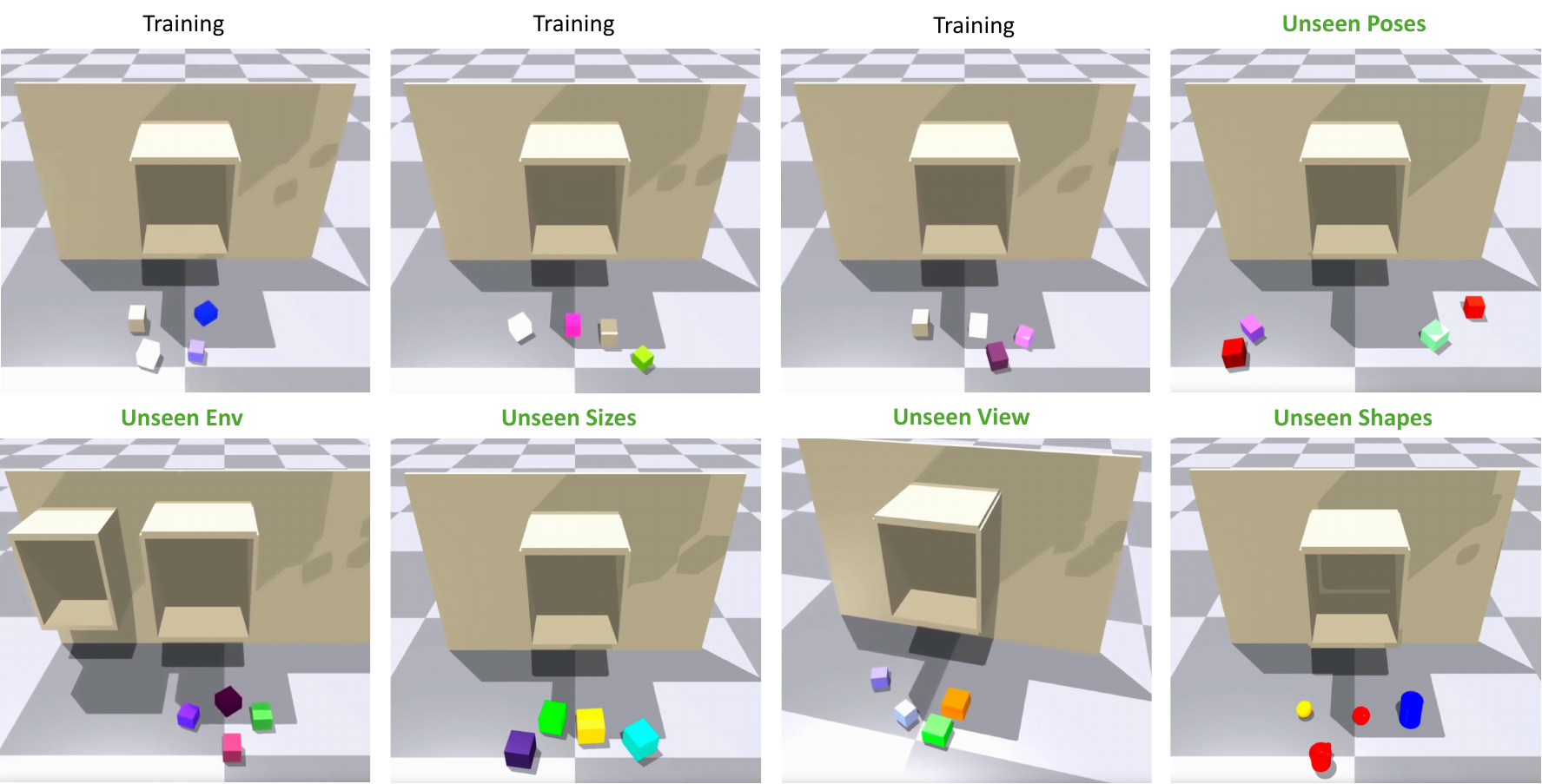}
    \vspace{5pt}
    \caption{Examples of our training dataset and some test dataset with unseen poses, environments, sizes, view, and shapes.  
    }
\label{fig:generalization}
\end{figure}

\begin{table*}[h]
\centering
\caption{Generalization to Unseen Scenarios}
\begin{tabular}{ p{5.8cm}p{3.8cm}p{3.8cm}}
 \hline
 Method & \cp{} & \eRDg{} \\
 \hline
 Unseen sizes of objects   & \textbf{58\%} & 33\%  \\
 Unseen camera view angles     & \textbf{61\%} & 42\%  \\
 Unseen environments     & \textbf{50\%} & 33\%  \\
 Unseen shapes of objects (YCB objects)     & \textbf{58\%} & 49\%  \\
 Unseen poses of objects     & \textbf{69\%} & 51\%  \\
\end{tabular}
\label{table:unseen}
\end{table*}

We show the extra simulation success rates of Points2Plans and the best-performing baseline to demonstrate the model’s generalization ability to unseen camera viewpoints, environments, and different sizes, shapes, and poses of objects. We run 100 trials per approach per generalization metric in the constrained packing task. 

From the results shown in the table.~\ref{table:unseen}, we find Points2Plans performs well when it generalizes to unseen scenarios and outperforms the baseline. Please refer to Fig.~\ref{fig:generalization} for the visualizations of comparisons between training datasets and test datasets with novel scenes.

\subsection{Generalization to Noisy Segmentation Masks}\label{sec:segmentation}
\begin{figure}[h]
    \centering
    \includegraphics[width=0.99\columnwidth]{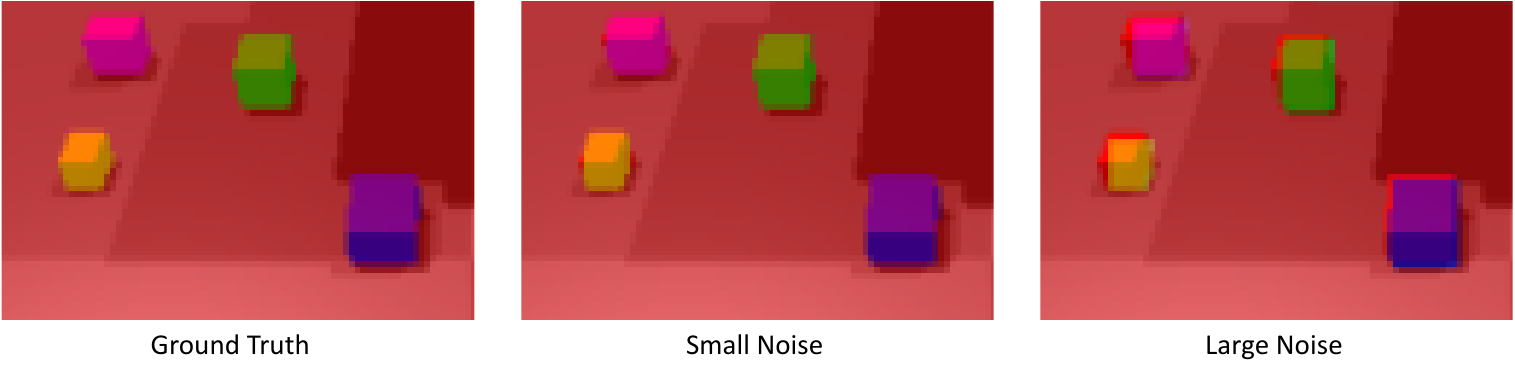}
    \caption{Examples of ground truth segmentation mask and noisy segmentation masks. 
    }
    \label{fig:noisy_masks}
\end{figure}

\begin{table*}[h]
\centering
\caption{Robustness to Noisy Segmentation Masks}
\begin{tabular}{ p{5.8cm}p{3.8cm}p{3.8cm}}
 \hline
 Method & \cp{} & \eRDg{} \\
 \hline
 No Noise   & \textbf{81\%} & 69\%  \\
 Small Noise     & \textbf{78\%} & 67\%  \\
 Large Noise     & \textbf{73\%} & 39\%  \\
\end{tabular}
\label{table:noise}
\end{table*}

We show our method’s and the best-performing baseline’s robustness to the noise in the segmentation mask. We use the erosion and dilation algorithm~\cite{gil2002efficient} to generate the noise for the segmentation masks. We use kernel size = 5 for generating small noise and kernel size = 10 for generating large noise to segmentation masks. We run 100 trials per approach per noise metric in the constrained packing task. 

From the results shown in the table.~\ref{table:noise}, we find Points2Plans performs well in the robustness to the noise for the segmentation mask while the baseline performs poorly, especially with large noise. 
Please refer to Fig.~\ref{fig:noisy_masks} for the details and the visualizations of noisy segmentation masks.

\subsection{Additional Related Work}\label{sec:relatedwork}
\textbf{Task and motion planning} solves long-horizon tasks through symbolic and geometric reasoning~\cite{toussaint2015logic, garrett2020pddlstream, garrett2021integrated, vu2024coast}.
Perception modules can be used to alleviate TAMP's assumption on full state observability~\cite{curtis2022long}.
Other works learn vision-based planning heuristics~\cite{driess2020deep1, driess2020deep2} or behavior policies~\cite{driess2021learning, dalal2023imitating} from pre-computed TAMP solutions.
We highlight two distinctions of our approach compared to TAMP: our system a) predicts symbolic effects instead of using predefined symbolic operators in e.g., PDDL~\cite{aeronautiques1998pddl} and b) plans in a latent space directly encoded from 3D partial-view point clouds.
Our approach is related to learning symbolic operators~\cite{ames2018learning, konidaris2018skills, silver2021learning, wang2021learning, silver2023predicate} and object dynamics~\cite{stap, chitnis2022learning} for long-horizon planning, but differs in the use of our \rd{} model, which captures both symbolic and geometric effects of actions in a shared latent space.

\subsection{Detailed Limitations}\label{sec:limitations}
\cp{} executes its plans in an open-loop fashion, i.e., without considering feedback from the environment.
Exploring closed-loop strategies for refining plans or correcting execution failures via replanning are possible points of extension.
Our relational dynamics model currently only predicts object positions when rolling out a plan, which is sufficient for tasks involving simple (e.g., symmetric) object geometries. 
Scaling \cp{} to more complex object geometries necessitates the prediction of their full pose.
We also assume access to a set of hand-designed manipulation primitives.
Interfacing with closed-loop policies~\cite{huang2023voxposer, chi2023diffusionpolicy} might allow \cp{} to solve tasks that require more fine-grained motions. 
Furthermore, our framework assumes a fixed set of predicates to learn skill effects.
Drawing from predicate learning techniques~\cite{shah2024reals, ahmetoglu2024discovering} could improve the generality of our framework, e.g., facilitating planning in an open-world setting. 
Finally, while we demonstrate faster planning with LLMs, their task planning performance degrades with longer tasks and more complex instructions~\cite{kambhampati2024llms}. 
Including more sophisticated LLM-based task planning strategies~\cite{lin2023text2motion} would improve the overall robustness of our planner.

\end{document}